\documentclass{article} % For LaTeX2e
\usepackage{iclr2023_conference,times}

\usepackage{amsmath}
\usepackage{amssymb}
\usepackage{hyperref}
\usepackage{url}
\usepackage{acro} % handles abbreviations and citations
\usepackage[nameinlink,capitalize]{cleveref} % provides \cref and \Cref
\usepackage{tabularx}       % tables that can fit the full width of column
\usepackage{multirow}
\usepackage{booktabs}
\usepackage{makecell} % line breaks within a table cell
\usepackage{subcaption} % newer and more maintained than subfig
\usepackage{chngcntr} % number appendix figures
\usepackage{xargs}          % Use more than one optional parameter in a new commands
\usepackage{mathtools} % for \coloneqq
\usepackage{xspace} % add space automatically after \newcommand
\usepackage{acro} % handles abbreviations and citations
\usepackage{microtype}      % microtypography
\usepackage{pifont} % gives us \ding http://ctan.org/pkg/pifont
\usepackage{enumitem} % reduce margin of enumerate items

\title{Factorized Fourier Neural Operators}

\author{Alasdair Tran$^1$
\And
Alexander Mathews $^1$
\And
Lexing Xie $^1$
\And
Cheng Soon Ong $^{1, 2}$ \\
\AND
\textnormal{$^1$ Australian National University}
\and
\textnormal{$^2$ Data61, CSIRO}
}

\definecolor{ruby}{rgb}{0.6, 0.1, 0.3}

\definecolor{navy}{rgb}{0.1, 0.1, 0.8}
\definecolor{olive}{rgb}{0.1, 0.6, 0.2}

\newcommand{\cmark}{\ding{51}}%

\newcommandx{\unsure}[2][1=]{\todo[linecolor=red,backgroundcolor=red!25,bordercolor=red,#1]{#2}}
\newcommandx{\change}[2][1=]{\todo[linecolor=blue,backgroundcolor=blue!25,bordercolor=blue,#1]{#2}}
\newcommandx{\infoneeded}[2][1=]{\todo[linecolor=olive,backgroundcolor=olive!25,bordercolor=olive,#1]{#2}}

\newcommand{\TorusZongyi}{{TorusLi}\xspace} %
\newcommand{\TorusKochkov}{{TorusKochkov}\xspace} %
\newcommand{\TorusV}{{TorusVis}\xspace} % TorusVis, TorusV, Vis
\newcommand{\TorusVF}{{TorusVisForce}\xspace} % TorusVisForce, TorusVisForcing, TorusVF, VisForcing
\newcommand{\Elasticity}{{Elasticity}\xspace}
\newcommand{\Plasticity}{{Plasticity}\xspace}
\newcommand{\Airfoil}{{Airfoil}\xspace}

\usepackage{adjustbox} % for rotating headers
\newcolumntype{R}[2]{%
    >{\adjustbox{angle=#1,lap=\width-(#2)}\bgroup}%
    l%
    <{\egroup}%
}
\newcommand*\rot{\multicolumn{1}{R{45}{1em}}}% no optional argument here, please!

\acsetup{
  single         = true ,
  list/sort      = true ,
  cite/cmd       = \citep ,
  cite/group = false ,
  cite/group/cmd = \citep,
}

\DeclareAcronym{FNO}{
  short = FNO,
  long = Fourier Neural Operator,
}

\DeclareAcronym{F-FNO}{
  short = F-FNO,
  long = Factorized Fourier Neural Operator,
}

\DeclareAcronym{DNS}{
  short = DNS,
  long = Direct Numerical Simulation,
}

\DeclareAcronym{PDE}{
  short = PDE,
  long = partial differential equation,
}

\DeclareAcronym{CFL}{
  short = CFL,
  long = Courant-Friedrichs-Lewy,
}

\DeclareAcronym{FEM}{
  short = FEM,
  long = finite element method,
}

\DeclareAcronym{FDM}{
  short = FDM,
  long = finite difference method,
}

\DeclareAcronym{FVM}{
  short = FVM,
  long = finite volume method,
}

\makeatletter
    \def\tagform@#1{\maketag@@@{\normalsize(#1)\@@italiccorr}}
\makeatother

\iclrfinalcopy % Uncomment for camera-ready version, but NOT for submission.

\begin{document}

\maketitle
% \fancyhead[L]{Under review}
\vspace{-0.5cm}

% !TEX root = ../main.tex

\begin{abstract}
We propose the \ac{F-FNO}, a learning-based approach for simulating \acfp{PDE}.
%Recent machine learning approach for solving \acp{PDE} include hybrid methods
%that replace parts of numerical solvers with learnt alternatives, or
%data-driven models with no knowledge of the underlying physical equation. The
%latter is arguably more flexible, but has not matched the traditional or
%hybrid solvers in terms of solution quality.
Starting from a recently proposed Fourier representation of flow fields, the
\ac{F-FNO} bridges the performance gap between pure machine learning approaches
to that of the best numerical or hybrid solvers. This is achieved with
new representations -- separable spectral layers and improved residual
connections -- and a combination of
%several
%insights that collectively have a significant effect -- the separable spectral
%representations; improved residual connections; and
training strategies such as the Markov assumption, Gaussian noise, and cosine
learning rate decay. On several challenging benchmark \acp{PDE} on regular
grids, structured meshes, and point clouds, the \ac{F-FNO} can scale to deeper
networks and outperform both the \acs{FNO} and the geo-FNO, reducing the error
by 83\% on the Navier-Stokes problem, 31\% on the elasticity problem, 57\% on
the airfoil flow problem, and 60\% on the plastic forging problem. Compared to
the state-of-the-art pseudo-spectral method, the \ac{F-FNO} can take a step
size that is an order of magnitude larger in time and achieve an order of
magnitude speedup to produce the same solution quality.

    %The \ac{FNO} is a learning-based method for efficiently simulating partial
    %differential equations. We propose the \ac{F-FNO} that allows much better
    %generalization with deeper networks. With a careful combination of the
    %Fourier factorization, a shared kernel integral operator across all layers,
    %the Markov property, and residual connections, the \ac{F-FNO} achieves a
    %six-fold reduction in error on the most turbulent setting of the
    %Navier-Stokes benchmark dataset, while using two orders of magnitude fewer
    %parameters than the original \ac{FNO}. Our method also uses an order of
    %magnitude less compute time than numerical solvers to achieve the same
    %correlation with the ground truths. Finally we show the \ac{F-FNO}'s
    %generalization ability by using same pretrained model to solve
    %Navier-Stokes equations with different viscosities and time-varying forces.

    % We also show that our model maintains an error rate of 2\% even when the
    % problem setting is extended to include additional contexts such as
    % viscosity and time-varying forces. This enables the same pretrained
    % neural network to model vastly different conditions.
\end{abstract}

\acresetall
% !TEX root = ../main.tex

\section{Introduction}
\label{sec:intro}

From modeling population dynamics to understanding the formation of stars,
\acp{PDE} permeate the world of science and engineering. For most real-world
problems, the lack of a closed-form solution requires using computationally
expensive numerical solvers, sometimes consuming millions of core hours and
terabytes of storage~\citep{Hosseini2016Direct}. Recently, machine learning
methods have been proposed to replace part~\citep{Kochkove2021Machine} or
all~\citep{Li2021Fourier} of a numerical solver.

Of particular interest are \acp{FNO}~\citep{Li2021Fourier}, which are neural
networks that can be trained end-to-end to learn a mapping between
infinite-dimensional function spaces. The \ac{FNO} can take a step size much
bigger than is allowed in numerical methods, can perform super-resolution, and
can be trained on many \acp{PDE} with the same underlying architecture.
% The \ac{FNO} is both efficient and
% generalizable. The efficiency of the \ac{FNO} comes from
% \textit{superresolution} (i.e., being able to model the finer details that a
% numerical solver would miss on the same resolution) and from \textit{leaping in
% time} (i.e., taking a step size much bigger than is allowed in numerical
% methods)
% Furthermore, since the \ac{FNO} does not encode specialized
% constraints, it can be be trained on many %any
% \acp{PDE} with the same % say things without negation
%out modifying the
% underlying architecture.
A more recent variant, %of \ac{FNO},
dubbed geo-FNO~\citep{Li2022FourierNO}, can handle irregular geometries such as
structured meshes and point clouds. However, this first generation of neural
operators suffers from stability issues. \citet{Lu2022ACA} find that the
performance of the \ac{FNO} deteriorates significantly on complex geometries
and noisy data. In our own experiments, we observe that both the \ac{FNO} and
the geo-FNO perform worse as we increase the network depth, eventually failing
to converge at 24 layers. Even at 4 layers, the error between the \ac{FNO} and
a numerical solver remains large (14\% error on the Kolmogorov flow).

In this paper, we propose the \ac{F-FNO} which contains an improved
representation layer for the operator, and a better set of training approaches.
By learning features in the Fourier space in each dimension independently, a
process called Fourier factorization, we are able to reduce the model
complexity by an order of magnitude and learn higher-dimensional problems such
as the 3D plastic forging problem. The F-FNO places residual connections after
activation, enabling our neural operator to benefit from a deeply stacked
network. Coupled with training techniques such as teacher forcing, enforcing
the Markov constraints, adding Gaussian noise to inputs, and using a cosine
learning rate scheduler, we are able to outperform the state of the art by a
large margin on three different \ac{PDE} systems and four different geometries.
On the Navier-Stokes (Kolmogorov flow) simulations on the torus, the \ac{F-FNO}
reduces the error by 83\% compared to the FNO, while still achieving an order
of magnitude speedup over the state-of-the-art pseudo-spectral method
(\cref{fig:torus_li_performance,fig:correlation}). On point clouds and
structured meshes, the \ac{F-FNO} outperforms the geo-FNO on both structural
mechanics and fluid dynamics PDEs, reducing the error by up to 60\%
(\cref{tab:meshes}).

Overall, we make the following three key contributions:

\begin{enumerate} %[leftmargin=0.5cm,itemsep=.1em] % noitemsep
      \item We propose a new representation, the \ac{F-FNO}, which consists of
            separable Fourier representation and improved residual connections,
            reducing the model complexity and allowing it to scale to deeper
            networks (\cref{fig:model,eqn:layer,eqn:separateFFT}).
            % and a shared kernel integral
            % operator across all layers to reduce the parameter count
            %from 100M to 1M
            % by two orders of magnitude and enable the network to grow to 24 layers on a single GPU.
            %(\cref{fig:param_count}).
      \item We show the importance of incorporating training techniques from
            the existing literature, such as Markov assumption, Gaussian noise,
            and cosine learning rate decay (\cref{fig:torus_li_performance});
            and investigate how well the operator can handle different input
            representations (\cref{fig:flexible-inputs}).

      \item We demonstrate \ac{F-FNO}'s strong performance in a variety of
            geometries and PDEs (\cref{fig:torus_li_performance,tab:meshes}).
            Code, datasets, and pre-trained models are
            available\footnote{\url{https://github.com/alasdairtran/fourierflow}}.
            % using meshes
            % and point clouds, the \ac{F-FNO} outperforms the geo-FNO by
            % a significant margin (\cref{tab:meshes}).
            % The \ac{F-FNO} can also
            % take additional contexts as input, allowing us to solve more
            % general PDEs with different parameters {\em
            % without the need to retrain} (\cref{fig:flexible-inputs}). Code and datasets will be released on
            % GitHub after publication.%\footnote{\url{https://anonymized}}
\end{enumerate}

\section{Related Work}
\label{sec:related}

Classical methods to solve \ac{PDE} systems include \acp{FEM}, \acp{FDM},
\acp{FVM}, and pseudo-spectral methods such as Crank-Nicholson and
Carpenter-Kennedy. In these methods, space is discretized, and a more accurate
simulation requires a finer discretization which increases the computational
cost. Traditionally, we would use simplified models for specific \acp{PDE},
such as Reynolds averaged Navier-Stokes~\citep{Alfonsi2009ReynoldsAveragedNE}
and large eddy simulation~\citep{Lesieur1996NewTI}, to reduce this cost. More
recently, machine learning offers an alternative approach to accelerate the
simulations. There are two main clusters of work: hybrid approaches and pure
machine learning approaches. Hybrid approaches replace parts of traditional
numerical solvers with learned alternatives but keep the components that impose
physical constraints such as conservation laws; while pure machine learning
approaches learn the time evolution of \acp{PDE} from data only.

\paragraph{Hybrid methods}

typically aim to speed up traditional numerical solvers by using lower
resolution grids~\citep{BarSinai2019Learning, Um2020Solver,
Kochkove2021Machine}, or by replacing computationally expensive parts of the
solver with learned alternatives~\cite{tompson2017accelerating,
obiols2020cfdnet}. \citet{BarSinai2019Learning} develop a data driven method
for discretizing \ac{PDE} solutions, allowing coarser grids to be used without
sacrificing detail. \citet{Kochkove2021Machine} design a technique specifically
for the Navier-Stokes equations that uses neural network-based interpolation to
calculate velocities between grid points rather than using the more traditional
polynomial interpolation. Their method leads to more accurate simulations while
at the same time achieving an 86-fold speed improvement over \ac{DNS}.
Similarly, \citet{tompson2017accelerating} employ a numerical solver and a
decomposition specific to the Navier-Stokes equations, but introduce a
convolutional neural network to infer the pressure map at each time step. While
these hybrid methods are effective when designed for specific equations, they are
not easily adaptable to other \ac{PDE} tasks.

An alternative approach, less specialized than most hybrid methods but
also less general than pure machine learning methods, is learned
correction~\citep{Um2020Solver, Kochkove2021Machine} which involves learning a
residual term to the output of a numerical step. That is, the time derivative
is now $\mathbf{u}_t = \mathbf{u}^*_t + LC(\mathbf{u}^*_t)$, where
$\mathbf{u}^*_t$ is the velocity field provided by a standard numerical solver
on a coarse grid, and $LC(\mathbf{u}^*_t)$ is a neural network that plays the
role of super-resolution of missing details.

%involves replacing part of the convection term in the
%Navier-Stokes equations with a neural network. In particular, to calculate the
%convective flux $\mathbf{u} \otimes \mathbf{u}$, we need to interpolate the
%velocities between the grid points. Instead of the traditionally-used
%polynomial interpolation which performs poorly as we lower the spatial
%resolution, \citet{Kochkove2021Machine} showed that learned
%interpolation~\citep{BarSinai2019Learning}, in which the interpolation
%coefficients are learned from data, leads to more accurate simulations while at
%the same time achieving an 80-fold speed improvement over \ac{DNS}.

\paragraph{Pure machine learning approaches}

eschew the numerical solver altogether and learn the field directly, i.e.,
$\mathbf{u}_t = \mathcal{G}(\mathbf{u}_{t-1})$, where $\mathcal{G}$ is dubbed a
neural operator. The operator can include graph neural
networks~\cite{Li2020MultipoleGN, Li2020NeuralOG}, low-rank
decomposition~\cite{Kovachki2021NeuralOL}, or Fourier
transforms~\citep{Li2021Fourier,Li2021MarkovNO}. Pure machine learning models
can also incorporate physical constraints, for example, by carefully designing
loss functions based on conservation laws~\citep{wandel2020learning}. They can
even be based on existing simulation methods such as the operator designed by
\citet{wang2020towards} that uses learned filters in both Reynolds-averaged
Navier-Stokes and Large Eddy Simulation before combining the predictions using
U-Net. However, machine learning methods need not incorporate such
constraints -- for example, \citet{kim2019deep} use a generative CNN model to
represent velocity fields in a low-dimensional latent space and a feedforward
neural network to advance the latent space to the next time point. Similarly,
\citet{bhattacharya2020model} use PCA to map from an infinite dimensional input
space into a latent space, on which a neural network operates before being
transformed to the output space. Our work is most closely related to the
Fourier transform-based approaches~\citep{Li2021Fourier,Li2022FourierNO} which
can efficiently model \acp{PDE} with zero-shot super-resolution but is not
specific to the Navier-Stokes equations.

%This allows some simulations to be run up to 700x faster than existing numerical solvers.

\paragraph{Fourier representations}

%Fourier transforms have gone through a long history in deep learning. For a
%while, frequency-domain representations fell out of popularity in favor of
%learned features~\citep{amodei2016deepspeech2}.

are popular in deep-learning due to the efficiency of convolution operators in the
frequency space, the $O(n \log n)$ time complexity of the fast Fourier
transform (FFT), and the ability to capture long-range dependencies. Two
notable examples of deep learning models that employ Fourier representations
are FNet~\citep{LeeThorp2021FNetMT} for encoding semantic relationships in text
classification and \ac{FNO}~\citep{Li2021Fourier} for flow simulation. In
learning mappings between function spaces, the \ac{FNO} outperforms graph-based
neural operators and other finite-dimensional operators such as U-Net. In
modeling chaotic systems, the \ac{FNO} has been shown to capture invariant
properties of chaotic systems~\citep{Li2021MarkovNO}. More generally,
\citet{Kovachki2021OnUA} prove that the \ac{FNO} can approximate any
continuous operator.

% !TEX root = ../main.tex

\section{The Factorized Fourier Neural Operator}
\label{sec:model}

\begin{figure*}[t]
  \centering
  \includegraphics[width=\linewidth]{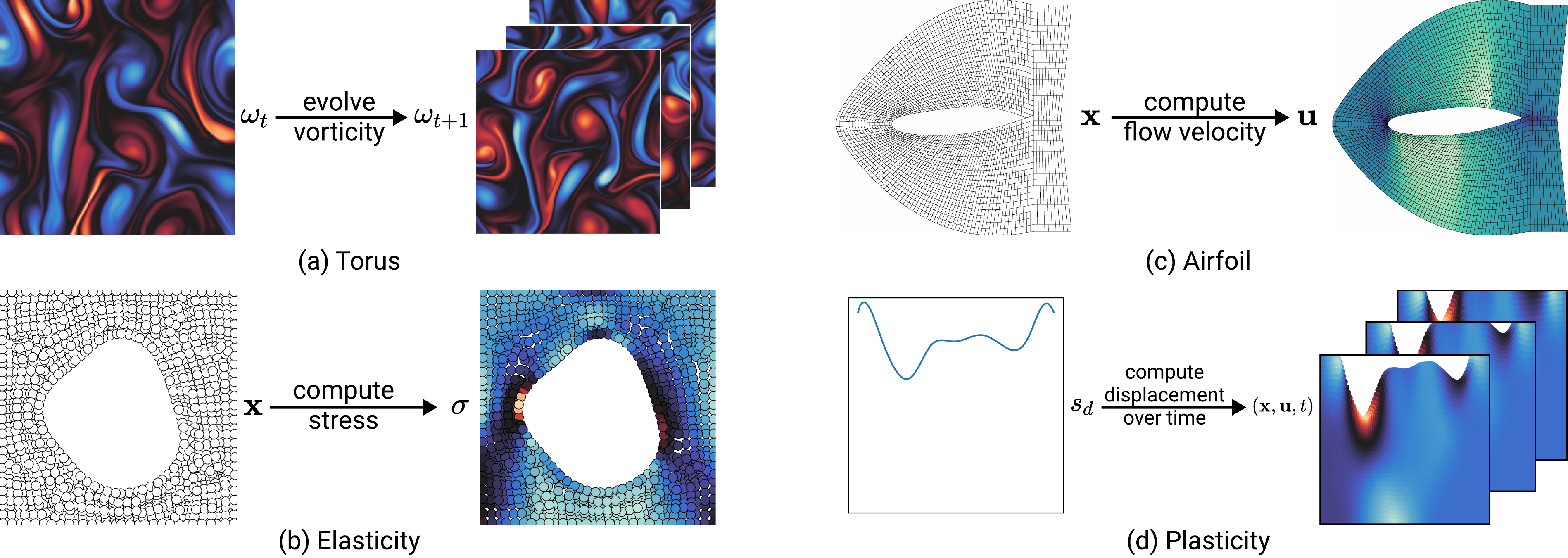}
  \centering{
    \phantomsubcaption\label{fig:viz_torus}
    \phantomsubcaption\label{fig:viz_elasticity}
    \phantomsubcaption\label{fig:viz_airfoil}
    \phantomsubcaption\label{fig:viz_plasticity}
  }
  \caption{An illustration of the input and output of different PDE problems.
  See the accompanying \cref{tab:datasets} for details.
  On the
  torus datasets (a), the operator learns to
  evolve the vorticity over time. On \Elasticity (b), the operator learns to
  predict the stress value on each point on a point cloud. On \Airfoil (c), the
  operator learns to predict the flow velocity on each mesh point. On
  \Plasticity (d), the operator learns the displacement of
  each mesh point given an initial boundary condition.}
  \label{fig:data-viz}
\end{figure*}

\paragraph{Solving \acp{PDE} with neural operators}

An operator $\mathcal{G} : \mathcal{A} \rightarrow \mathcal{U}$ is a mapping
between two infinite-dimensional function spaces $\mathcal{A}$ and
$\mathcal{U}$. Exactly what these function spaces represent depends on the
problem. In general, solving a \ac{PDE} involves finding a solution $u \in
\mathcal{U}$ given some input parameter $a \in \mathcal{A}$, and we would train
a neural operator to learn the mapping $a \mapsto u$. Consider the vorticity
formulation of the 2D Navier-Stokes equations,
\begin{align}
  \frac{\partial \omega}{\partial t} + \mathbf{u} \cdot \nabla \omega
    = \nu \nabla^2 \omega + f
\qquad \qquad
\nabla \cdot \mathbf{u} = 0 \label{eqn:ns-vorticity}
\end{align}
where $\mathbf{u}$ is the velocity field, $\omega$ is the vorticity, and $f$ is
the external forcing function.
% \cref{eqn:ns-vorticity} can be obtained by taking the curl of
% \cref{eqn:ns-velocity}. It is often easier to work with the vorticity
% formulation since the neural network only needs to output scalar numbers
% (i.e., only one component in the Cartesian coordinates of the vorticity field
% in 2D is non-zero).
These are the governing equations for the torus
datasets~(\cref{fig:viz_torus}). The neural operator would learn to evolve this
field from one time step to the next: $\omega_t \mapsto \omega_{t+1}$. Or
consider the equation for a solid body in structural mechanics,
\begin{align}
  \rho \dfrac{\partial^2 \mathbf{u}}{\partial t^2} + \nabla \cdot \mathbf{\sigma} = 0,
\end{align}
where $\rho$ is the mass density, $\mathbf{u}$ is the displacement vector and
$\mathbf{\sigma}$ is the stress tensor. \Elasticity~(\cref{fig:viz_elasticity})
and \Plasticity~(\cref{fig:viz_plasticity}) are both governed by this equation.
In \Plasticity, we would learn to map the initial boundary condition $s_d : [0,
L] \rightarrow \mathbb{R}$ to the grid position $\mathbf{x}$ and displacement
of each grid point over time: $s_d \mapsto (\mathbf{x}, \mathbf{u}, t)$. In
\Elasticity, we are instead interested in predicting the stress value for each
point: $\mathbf{x} \mapsto \sigma$. Finally consider the Euler equations to
model the airflow around an aircraft wing~(\cref{fig:viz_airfoil}):
\begin{align}
    \dfrac{\partial \rho}{\partial t} + \nabla \cdot (\rho \mathbf{u}) = 0
    \qquad
    \dfrac{\partial\rho\mathbf{u}}{\partial t}
      + \nabla \cdot (\rho \mathbf{u} \otimes \mathbf{u} + p \mathbb{I}) = 0
    \qquad
    \dfrac{\partial E}{\partial t} + \nabla \cdot ((E + p) \mathbf{u}) = 0
    \label{eqn:euler}
\end{align}
where $\rho$ is the fluid mass density, $p$ is the pressure, $\mathbf{u}$ is
the velocity vector, and $E$ is the energy. Here the operator would learn to
map each grid point to the velocity field at equilibrium: $\mathbf{x} \mapsto
\mathbf{u}$.

\begin{figure*}[t]
    \centering
    \includegraphics[width=0.95\linewidth]{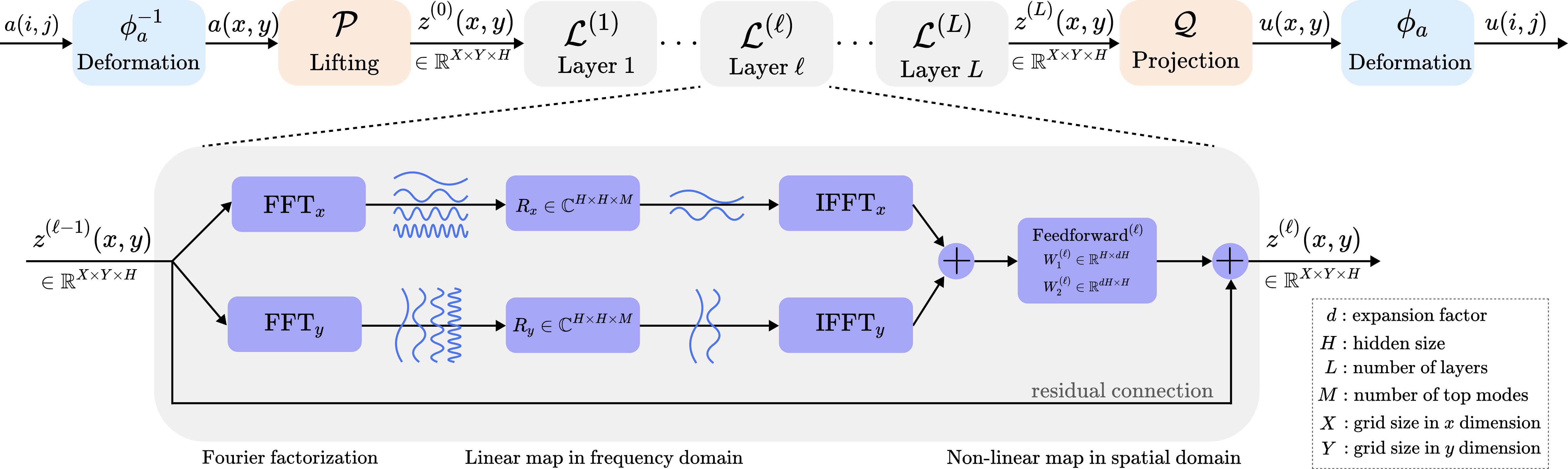}
    \caption{The architecture of the \acf{F-FNO} for a 2D problem. The
    iterative process (\cref{eqn:process}) is shown at the top, in which the
    input function $a(i,j)$ is first deformed from an irregular space into a
    uniform space $a(x,y)$, and is then fed through a series of operator layers
    $\mathcal{L}$ in order to produce the output function $u(i,j)$. A zoomed-in
    operator layer (\cref{eqn:layer}) is shown at the bottom which shows how we
    process each spatial dimension independently in the Fourier space, before
    merging them together again in the physical space.}
    \label{fig:model}
\end{figure*}

\paragraph{Original \ac{FNO} and geo-FNO architectures}

% The task of a \ac{PDE} solver is to evolve the field from one time step to the
% next: $\omega_t \rightarrow \omega_{t+1}$.
% These fields are defined over a bounded spatial domain, $\omega_t : D
% \rightarrow \mathbb{R}$ with $D \subset \mathbb{R}^2$, and they
% live in a function space, $\omega_t \in \Omega(D; \mathbb{R})$.
% Under this formulation, we can frame
% our problem as learning a neural operator between infinite-dimensional function
% spaces. The operator
% $\mathcal{G} : \Omega \rightarrow \Omega$ would then learn to map the field
% $\omega_t$ at some time step $t$ to the next time step $t+1$.

Motivated by the kernel formulation of the solution to linear \acp{PDE} using
Green's functions, \citet{Li2020NeuralOG,Li2022FourierNO} propose an iterative
approach to map input function $a$ to output function $u$,
\begin{align}
    u = \mathcal{G}(a) &=
        (\phi \circ \mathcal{Q} \circ \mathcal{L}^{(L)} \circ
        \dots \circ \mathcal{L}^{(1)} \circ
        \mathcal{P} \circ \phi^{-1}) (a), \label{eqn:process}
\end{align}
where $\circ$ indicates function composition, $L$ is the number of
layers/iterations, $\mathcal{P}$ is the lifting operator that maps the input to
the first latent representation $z^{(0)}$, $\mathcal{L}^{(\ell)}$ is the
$\ell$'th non-linear operator layer, and $\mathcal{Q}$ is the projection
operator that maps the last latent representation $z^{(L)}$ to the output. On
irregular geometries such as point clouds, we additionally define a coordinate
map $\phi$, parameterized by a small neural network and learned end-to-end,
that deforms the physical space of irregular geometry into a regular
computational space. The architecture without this coordinate map is called
\ac{FNO}, while the one with the coordinate map is called geo-FNO.
% we would also like to find a diffeomorphic
% deformation between the input domain (e.g., the point cloud) and the unit
% torus. This would allow us to continue using \cref{eqn:process}, but with an
% extra deformation function in the input and output. Following
% \citet{Li2022FourierNO}, we use the geometric Fourier Transform in the first
% layer, and the inverse geometric Fourier Transform in the last layer. Both of
% these transforms involve defining a coordinate map, parameterized by a small
% neural network and learned end-to-end, that deforms the physical space of
% irregular geometry into a regular computational space which the standard
% Fourier transform can understand.
% So far, we have formulated the \ac{F-FNO} in the context of the 2D
% Navier-Stokes equations, where we learn an operator that maps the vorticity
% field from one time step to the next. But neural operators can solve more
% general problems. Consider the equation for a solid body in structural
% mechanics,
% $
%   \rho^s \dfrac{\partial^2 \mathbf{u}}{\partial t^2} + \nabla \cdot \mathbf{\sigma} = 0
% $,
% where $p^s$ is the mass density, $\mathbf{u}$ is the displacement vector and
% $\mathbf{\sigma}$ is the stress tensor. The \Elasticity dataset, governed by
% this equation and described in \cref{sec:datasets}, consists of point clouds,
% not regular grids. Our neural operator would then need to learn to map a point
% cloud to its corresponding stress value.
\cref{fig:model} (top)
contains a schematic diagram of this iterative process.

Originally, \citet{Li2021Fourier} formulate each operator layer as
\begin{align}
    \mathcal{L}^{(\ell)}\Big( z^{(\ell)} \Big) &= \sigma \Big(
        W^{(\ell)} z^{(\ell)} + b^{(\ell)} +
        \mathcal{K}^{(\ell)}(z^{(\ell)})
    \Big), \label{eqn:operator_layer}
\end{align}
where $\sigma : \mathbb{R} \rightarrow \mathbb{R}$ is a point-wise non-linear
activation function, $W^{(\ell)} z^{(\ell)} + b^{(\ell)}$ is an affine
point-wise map in the physical space, and $\mathcal{K}^{(\ell)}$ is a kernel
integral operator using the Fourier transform,
\begin{align}
    \mathcal{K}^{(\ell)}\Big( z^{(\ell)} \Big)  &= \text{IFFT}
        \Big(
            R^{(\ell)} \cdot \text{FFT}(z)
        \Big)  \label{eqn:kernel_operator}
\end{align}
The Fourier-domain weight matrices $\{R^{(\ell)} \mid \ell \in \{1, 2, \dots, L
\}\}$ take up most of the model size, requiring $O(LH^2M^D)$ parameters, where
$H$ is the hidden size, $M$ is the number of top Fourier modes being kept, and
$D$ is the problem dimension. Furthermore, the constant value for $M$ and the
affine point-wise map allow the \ac{FNO} to be resolution-independent.

\paragraph{Our improved F-FNO architecture}
% We consider a more general setting in which the operator $\mathcal{G}$ maps
% the vorticity field $a$ along with some contexts such as the viscosity $\nu$
% and the forcing function $f_t$, that is,
% \begin{align}
%     \omega_{t+1} = \mathcal{G}(\omega_t \mid \nu, f_t)
% \end{align}
We propose changing the operator layer in \cref{eqn:operator_layer} to:
\begin{align}
    \mathcal{L}^{(\ell)} \Big( z^{(\ell)} \Big) &=
        z^{(\ell)}  + \sigma \Big[
            W_2^{(\ell)} \sigma \big(W_1^{(\ell)}
            \mathcal{K}^{(\ell)}\big(z^{(\ell)}\big) +
            b^{(\ell)}_1 \big) + b^{(\ell)}_2
        \Big] \label{eqn:layer}
\end{align}
Note that
%unlike the original \ac{FNO} and geo-FNO,
we apply the residual
connection ($z^{(\ell)}$ term) \textit{after} the non-linearity to preserve
more of the layer input. We also use a two-layer feedforward, inspired
by the feedforward design used in transformers~\citep{Vaswani2017Attention}.
%Our kernel integral operator $\mathcal{K}$ is shared
%across all layers.
More importantly, we factorize the Fourier transforms over the problem dimensions,
modifying \cref{eqn:kernel_operator} to
\begin{align}
    \mathcal{K}^{(\ell)}\Big( z^{(\ell)} \Big)  &= \sum_{d \in D} \bigg[ \text{IFFT}
        \Big(
            R_d^{(\ell)} \cdot \text{FFT}_d(z^{(\ell)})
        \Big) \bigg]
        \label{eqn:separateFFT}
\end{align}
\noindent
% Our factorization pulls an exponent down from the model complexity --
The seemingly small change from $R^{(\ell)}$ to $R_d^{(\ell)}$ in the Fourier operator reduces the number of parameters to $O(LH^2MD)$.
%the \ac{F-FNO} now only requires $O(LH^2MD)$ parameters, which
This is particularly useful when solving higher-dimensional problems such as 3D
plastic forging (\cref{fig:viz_plasticity}). The combination of the factorized
transforms and residual connections allows the operator to converge in deep
networks while continuing to improve performance
(\cref{fig:torus_li_performance}). It is also possible to share the weight
matrices $R_d$ between the layers, which further reduces the parameters to
$O(H^2MD)$.
% \noindent Our approach uses two weight matrices, $R_x$ and $R_y$, that together
% have only $O(H^2M)$ parameters, irrespective of the number of layers.
\cref{fig:model} (bottom) provides an overview of an \ac{F-FNO} operator layer.

Furthermore, the \ac{F-FNO} is highly flexible in its input representation,
%The contextual information that the \ac{F-FNO} takes can be
which means anything that is relevant to the evolution of the field can be an
input, such as viscosity or external forcing functions for the torus.
%For example, if
%we know that an external force is acting on the field, we can simply add that
%force field as an additional input channel. In our experiments, we will show
%that each dataset makes use of a different set of contexts.
This flexibility also allows the \ac{F-FNO} to be easily generalized to
different \acp{PDE}.

% We will show in the next section that the factorization and weight sharing in
% $\mathcal{K}$, along with the residual connection in $\mathcal{L}$ and other
% deep learning techniques, allow the \ac{F-FNO} to vastly outperform the state of
% the art.

% Our operators remain resolution-independent, with all the weight matrices
% ($R_x$, $R_y$, $W^{\ell}_i$) not depending on the input grid size $N$.

\paragraph{Training techniques to learn neural operators}
We find that a combination of deep learning techniques are very important for
the FNO to perform well, most of which were overlooked in
\citet{Li2021Fourier}'s original implementation. The first is \textit{enforcing
the first-order Markov property}. We find \citet{Li2021Fourier}'s use of the
last 10 time steps as inputs to the neural operator to be unnecessary. Instead,
it is sufficient to feed information only from the current step, just like a
numerical solver. Unlike prior works~\citep{Li2021Fourier,
Kochkove2021Machine}, we do not unroll the model during training but instead
use the \textit{teacher forcing} technique which is often seen in time series
and language modeling. In teacher forcing, we use the ground truth as the input
to the neural operator. Finally during training, we find it useful to normalize
the inputs and add a small amount of Gaussian noise, similar to how
\citet{SanchezGonzalez2020LearningTS} train their graph networks. Coupled with
cosine learning rate decay, we are able to make the training process of neural
operators more stable. Ablation studies for the new representation and training
techniques can be found in \cref{fig:torus_li_performance}.
% Empirical results on the above three claims are in~\cref{sec:expt}.

% \paragraph{Connection with pseudo-spectral methods}

% Certain operations such as differentiation can be performed more efficiently by
% first taking the Fourier transform. Pseudo-spectral methods take advantage of
% this and compute the gradient $\nabla \omega$ and the Laplacian $\nabla^2
% \omega$ in the wavenumber space, while the divergent $\mathbf{u} \cdot \nabla$
% is done in the physical space. This back and forth between the two spaces can
% also be observed in the \ac{F-FNO}
% (see~\cref{eqn:layer,eqn:separateFFT}), although we let the
% network decide which features to learn in each space.

% !TEX root = ../main.tex

\section{Datasets and evaluation settings}
\label{sec:datasets}

\paragraph{PDEs on regular grids}

%We test the \ac{F-FNO} on seven \ac{PDE} datasets, covering three types of geometries -- regular grids, point clouds, and structured meshes (\cref{fig:data-viz}).
The four Torus datasets on regular grids (\TorusZongyi, \TorusKochkov, \TorusV, and \TorusVF, summarized in \cref{tab:datasets}) are
simulations based on Kolmogorov flows
which have been extensively studied in the
literature~\citep{Chandler2013Invariant}. In particular, they model turbulent
flows on the surface of a 3D torus (i.e., a 2D grid with periodic boundary
conditions). \TorusZongyi is publicly released by \citet{Li2021Fourier} and is
used to benchmark our model against the original \ac{FNO}. The ground truths
are assumed to be simulations generated by the pseudo-spectral Crank-Nicholson
second-order method on 64x64 grids. All trajectories have a constant viscosity
$\nu = 10^{-5}$ ($\text{Re} = 2000$), use the same constant forcing function,
$
  f(x, y) = 0.1 [
    \sin(2\pi (x + y)) +
    \cos(2\pi (x + y))
  ],
$
and differ only in the initial field.

\begin{table*}[t]
    \caption{An overview of the datasets and the corresponding task.}

    \label{tab:datasets}
    \begin{center}\small{
    \begin{tabularx}{\textwidth}{Xlclll}
      \toprule
      Dataset & Geometry & Dim. & Problem & Input & Output\\
      \midrule
      \TorusZongyi & regular grid & 2D & Kolmogorov flow & $\omega_t$ &  $\omega_{t+1}$\\
      \TorusKochkov & regular grid & 2D &  Kolmogorov flow & $\omega_t$ &  $\omega_{t+1}$ \\
      \TorusV & regular grid & 2D &  Kolmogorov flow & $\omega_t$ and $\nu$ &  $\omega_{t+1}$ \\
      \TorusVF & regular grid & 2D &  Kolmogorov flow & $\omega_t$ and $\nu$ and $f_t$ &  $\omega_{t+1}$ \\
      \Elasticity & point cloud & 2D &  hyper-elastic material
      & point cloud & stress \\
      \Airfoil & structured mesh & 2D & transonic flow & mesh grid & velocity \\
      \Plasticity & structured mesh & 3D &  plastic forging & boundary condition & displacement \\
      \bottomrule
    \end{tabularx}}
  \end{center}
  \end{table*}

Using the same Crank-Nicolson numerical solver, we generate two further
datasets, called \TorusV and \TorusVF, to test the generalization of the
\ac{F-FNO} across Navier-Stokes tasks with different viscosities and forcing
functions. In particular, for each trajectory, we vary the viscosity between
$10^{-4}$ and $10^{-5}$, and set the forcing function to
{\begin{equation}
  \begin{gathered}
  f(t, x, y) = 0.1 \sum_{p=1}^2 \sum_{i=0}^1 \sum_{j=0}^1 \Big[
    \alpha_{pij} \sin\big( 2\pi p (i x + j y) + \delta t \big) +
    \beta_{pij} \cos\big( 2\pi p (i x + j y) + \delta t \big)
  \Big],
  \end{gathered}
\end{equation}}%
where the amplitudes $\alpha_{pij}$ and $\beta_{pij}$ are sampled from the standard
uniform distribution. Furthermore, $\delta$ is set to 0 in \TorusV, making the
forcing function constant across time; while it is set to 0.2 in \TorusVF,
giving us a time-varying force.

Finally, we regenerate \TorusKochkov (\cref{fig:viz_torus}) using the same
settings provided by \citet{Kochkove2021Machine} but with different initial
conditions from the original paper (since the authors did not release the full
dataset). Here the ground truths are obtained from simulations on 2048x2048
grids using the pseudo-spectral Carpenter-Kennedy fourth-order method. The
full-scale simulations are then downsampled to smaller grid sizes, allowing us
to study the Pareto frontier of the speed vs accuracy space
(see~\cref{fig:pareto}). \TorusKochkov uses a fixed viscosity of $0.001$ and a
constant forcing function
$
\mathbf{f} = 4\cos(4y){\hat x} - 0.1  \mathbf{u}
$, but on the bigger domain of $[0, 2\pi]$. Furthermore, we generate only 32
training trajectories to test how well the \ac{F-FNO} can learn on a low-data
regime.

\paragraph{PDEs on irregular geometries} The \Elasticity, \Airfoil, and
\Plasticity datasets (final three rows in \cref{tab:datasets}) are taken from
\citet{Li2022FourierNO}. \Elasticity is a point cloud dataset modeling the
incompressible Rivlin-Saunders material \citep{Pascon2019LargeDA}. Each sample
is a unit cell with a void in the center of arbitrary shape
(\cref{fig:viz_elasticity}). The task is to map each cloud point to its stress
value. \Airfoil models the transonic flow over an airfoil, shown as the white
center in \cref{fig:viz_airfoil}. The neural operator would then learn to map
each mesh location to its Mach number. Finally, \Plasticity models the plastic
forging problem, in which a die, parameterized by an arbitrary function and
traveling at a constant speed, hits a block material from above
(\cref{fig:viz_plasticity}). Here the task is to map the shape of the die to
the $101 \times 31$ structured mesh over 20 time steps. Note that \Plasticity
expects a 3D output, with two spatial dimensions and one time dimension.

\paragraph{Training details}
For experiments involving the original \ac{FNO}, FNO-TF (with teaching
forcing), FNO-M (with the Markov assumption), and FNO-N (with improved
residuals), we use the same training procedure as \citet{Li2021Fourier}. For
our own models, we train for 100,000 steps on the regular grid datasets and for
200 epochs for the irregular geometry datasets, warming up the learning rate to
$2.5 \times 10^{-3}$ for the first 500 steps and then decaying it using the
cosine function~\citep{Loshchilov2017SGDR}. We use ReLU as our non-linear
activation function, clip the gradient value at 0.1, and use the Adam
optimizer~\citep{Kingma2015Adam} with $\beta_1 = 0.9$, $\beta_2 = 0.999$,
$\epsilon= 10^{-8}$. The weight decay factor is set to $10^{-4}$ and is
decoupled from the learning rate~\citep{Loshchilov2018Decoupled}. In each
operator layer on the torus datasets, we always throw away half of the Fourier
modes (e.g., on a 64x64 grid, we keep only the top 16 modes). Models are
implemented in PyTorch~\citep{Paszke2017Automatic} and trained on a single
Titan V GPU.

\paragraph{Evaluation metrics}
We use the normalized mean squared error
as the loss function, defined as
$$
  \text{N-MSE} =
    \frac{1}{B} \sum_{i=1}^{B}
    \frac{\lVert \hat{\omega}_i - \omega \rVert_2}
         {\lVert \omega \rVert_2},
$$
where $\lVert \cdot \rVert_2$ is the 2-norm, $B$ is the batch size, and
$\hat{\omega}$ is the prediction of the ground truth $\omega$.

In addition to comparing the N-MSE directly, for \TorusKochkov, we also compute the
vorticity correlation, defined as
$$
\rho(\omega, \hat{\omega})
= \sum_{i}\sum_{j} \frac{ \omega_{ij} }{\lVert \omega \rVert_2}
\frac{ \hat{\omega}_{ij} }{\lVert \hat{\omega} \rVert_2},
$$
and from which we measure the time until this correlation drops below 95\%.
To be consistent with prior work, we use the N-MSE to compare the \ac{F-FNO}
against the FNO and geo-FNO \citet{Li2021Fourier, Li2022FourierNO}, and the vorticity correlation to compare
against \citet{Kochkove2021Machine}'s work.

\section{Results for Naiver-Stokes on a torus}
\label{sec:results-torus}

\paragraph{Comparison against \ac{FNO}}

\begin{figure*}[t]
  \centering
  \includegraphics[width=\linewidth]{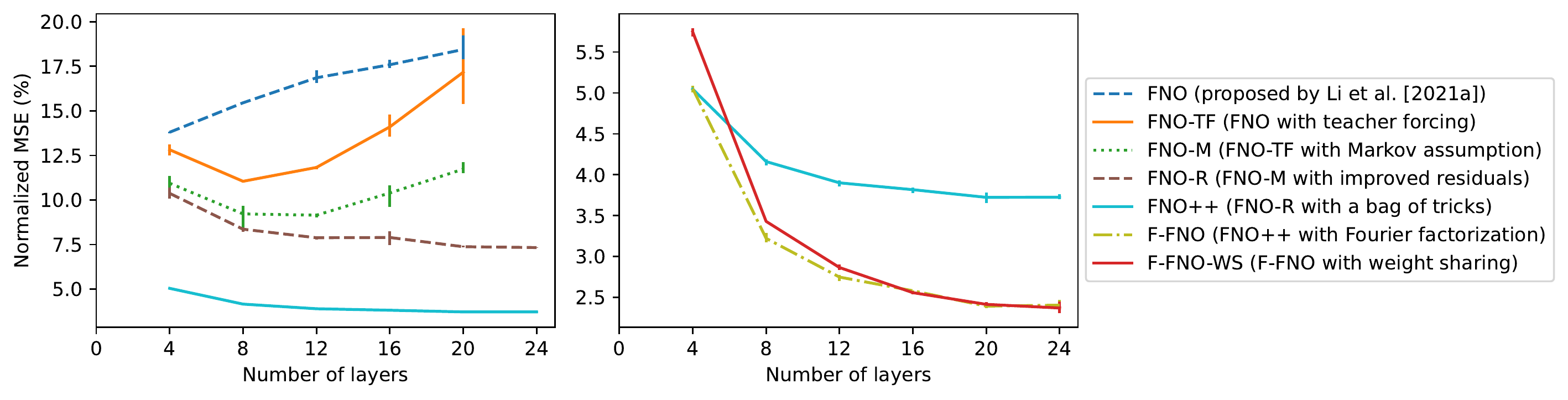}
  \
  \captionsetup[subfigure]{position=bottom,skip=2pt}
  \parbox{0.4\linewidth}{\subcaption{} \label{fig:mse_zongyi}}
  \parbox{0.3\linewidth}{\subcaption{} \label{fig:mse_ffno}}\hfill
  \
  \caption{Performance (lower is better)
           on \TorusZongyi,
           with error bars showing the min and max values over three trials.
           We show the original FNO~\citep{Li2021Fourier}, along
           with variants that use: teacher forcing, Markov assumption,
           improved residuals, a bag of tricks, Fourier factorization,
           and weight sharing. Note that F-FNO
           and F-FNO-WS are presented on a separate plot (b) to make
           visualizing the improvement easier (if shown in (a), F-FNO and F-FNO-WS would just be a straight line).
          %  In (c), we show
          %  the error (i.e., how much we diverge from \ac{DNS})
          %  as we run 10 seconds of simulation using a four-layer network.
%  Error bars show the min and
%  max values over three trials. % (most are too small to see).
%   \change{Can we use viridis and friends?}
  }
  \label{fig:torus_li_performance}
\end{figure*}

The performance on \TorusZongyi is plotted in
\cref{fig:torus_li_performance}, with the raw numbers shown
in~\cref{tab:results_torus_const}. We note that our method \ac{F-FNO} is
substantially more accurate than the \ac{FNO} regardless of network depth, when
judged by N-MSE. The \ac{F-FNO} uses fewer
parameters than the \ac{FNO}, has a similar training time, but generally has a
longer inference time. Even so, the inference
time for the \ac{F-FNO} is still up to two orders of magnitude shorter than for
the Crank-Nicolson numerical solver. % (\cref{fig:complexity} (c)).

In contrast to our method, \citet{Li2021Fourier} do not use teacher forcing
during training. Instead they use the previous 10 steps as input to predict the next 10 steps
incrementally (by using each predicted value as the input to the next step). We
find that the teacher forcing strategy (FNO-TF, orange line), in which we
always use the ground truth from the previous time step as input during
training, leads to a smaller N-MSE when the number of layers is less than 24.
Furthermore, enforcing the first-order Markov property (FNO-M, dotted green
line), where only one step of history is used, further improves the performance
over \ac{FNO} and FNO-TF. Including two or more steps of history does not
improve the results.

\begin{figure*}[t]
  \centering
  \includegraphics[width=\linewidth]{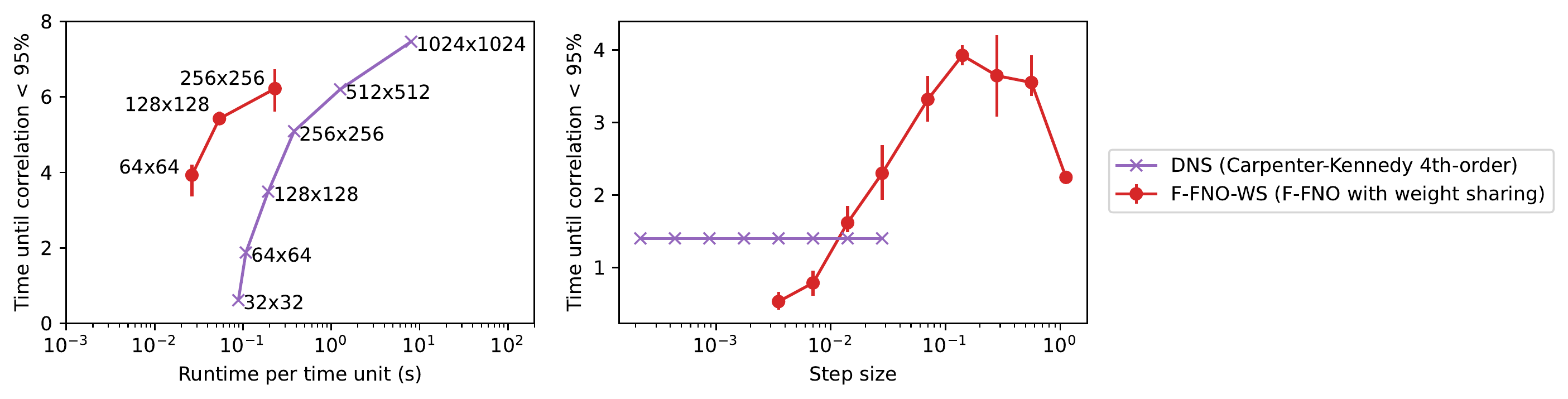}
  \
  \captionsetup[subfigure]{position=bottom,skip=2pt}
  \parbox{0.4\linewidth}{\subcaption{} \label{fig:pareto}}
  \parbox{0.3\linewidth}{\subcaption{} \label{fig:corr_stepsize}}\hfill
  \
  \caption{Performance of \ac{F-FNO} on \TorusKochkov. In (a), we plot the time
           until the correlation with the ground truths in the test set
           drops below 95\% on the
           y-axis, against the time it takes to run one second of simulation on
           the x-axis. In (b), we show how, on the validation set of
           \TorusKochkov, given a fixed spatial resolution of
           64x64, changing the step size has no effect on the numerical solver;
           however there is an optimal step size for the \ac{F-FNO} at around 0.2.}
  \label{fig:correlation}
\end{figure*}

% \begin{table}[t]
%   \caption{Performance on point clouds (\Elasticity) and structured meshes
%   (\Airfoil and \Plasticity). The N-MSE is accompanied by the standard
%   deviation from three trials.}

%   \label{tab:meshes-shared}
%   \begin{center}
%     \small{
%   \begin{tabularx}{\columnwidth}{Xrrrrrr}
%     \toprule
%        \multirow{2}{*}{No. of layers}
%      & \multicolumn{2}{c}{\Elasticity}
%      & \multicolumn{2}{c}{\Airfoil}
%      & \multicolumn{2}{c}{\Plasticity} \\
%      & geo-FNO & F-FNO & geo-FNO & F-FNO & geo-FNO & F-FNO \\
%     \midrule
%     4 layer & $2.5 \pm 0.1$ & $3.6 \pm 1.6$ & $1.9 \pm 0.4$ & $1.0 \pm 0.1$ & $0.74 \pm 0.01$ & $0.58 \pm 0.01$ \\
% 8 layer & $3.3 \pm 1.3$ & $2.2 \pm 0.0$ & $1.4 \pm 0.5$ & $0.7 \pm 0.0$ & $0.57 \pm 0.04$ & $0.50 \pm 0.02$ \\
% 12 layer & $16.8 \pm 0.7$ & $2.1 \pm 0.0$ & $4.1 \pm 4.4$ & $0.7 \pm 0.0$ & $0.45 \pm 0.03$ &  -  \\
% 16 layer & $16.3 \pm 0.4$ & $2.1 \pm 0.0$ & $16.2 \pm 0.0$ & $0.7 \pm 0.0$ & $34.56 \pm 24.08$ &  -  \\
% 20 layer & $16.0 \pm 0.7$ & $2.0 \pm 0.0$ & $16.2 \pm 0.0$ & $0.6 \pm 0.0$ &  -  &  -  \\
% 24 layer & $15.9 \pm 0.5$ &  -  & $16.2 \pm 0.0$ & $0.7 \pm 0.0$ &  -  &  -  \\
%     \bottomrule
%   \end{tabularx}}

% \end{center}
% \end{table}

The models FNO, FNO-TF, and FNO-M do not scale with network depth, as seen by
the increase in the N-MSE with network depth. These models even diverge during
training when 24 layers are used. FNO-R, with the residual connections placed
after the non-linearity, does not suffer from this problem and can finally
converge at 24 layers. FNO++ further improves the performance, as a result of a
careful combination of: normalizing the inputs, adding Gaussian noise to the
training inputs, and using cosine learning rate decay. In particular, we find that
adding a small amount of Gaussian noise to the normalized inputs helps to
stabilize training. Without the noise, the validation loss at the early stage
of training can explode.% as we unroll several steps into the future.

%we move the residual connection outside
%the non-linear activation to ease the gradient flow, and couple it with other
%deep learning techniques such as normalizing the inputs to have zero mean and
%unit variance, adding Gaussian noise to training inputs, and using a cosine
%learning rate scheduler with warmup (FNO++, cyan line).

Finally, if we use Fourier factorization (F-FNO, yellow dashed line), the error
drops by an additional 35\% ($3.73\% \rightarrow 2.41\%$) at 24 layers
(\cref{fig:mse_ffno}), while the parameter count is reduced by an order of
magnitude. %(\cref{fig:param_count}).
Sharing the weights in the Fourier domain (F-FNO-WS, red line) makes little
difference to the performance especially at deep layers, but it does reduce the
parameter count by another order of magnitude to 1M (see
\cref{fig:complexity,tab:results_torus_const}).
% Finally, even though our models do take longer to do
% inference, the deepest network is still an order of magnitude faster than the
% Crank-Nicholson numerical simulator (\cref{fig:inference_time}).

\paragraph{Trade-off between speed and accuracy}

From \cref{fig:pareto}, we observe that our method \ac{F-FNO} only needs 64x64
input grids to reach a similar performance to a 128x128 grid solved with
\ac{DNS}. At the same time, the \ac{F-FNO} also achieves an order of magnitude
speedup. While the highly specialized hybrid method introduced by
\citet{Kochkove2021Machine} can achieve a speedup closer to two orders of
magnitude over \ac{DNS}, the \ac{F-FNO} takes a much more flexible approach and
thus can be more easily adapted to other \acp{PDE} and geometries.

The improved accuracy of the \ac{F-FNO} over \ac{DNS} when both methods are
using the same spatial resolution can be seen graphically in \cref{fig:flows}.
In this example, the \ac{F-FNO} on a 128x128 grid produces a vorticity field
that is visually closer to the ground truth than \ac{DNS} running on the same
grid size. This is also supported by comparing the time until correlation falls
below $95\%$ in \cref{fig:pareto}.

\paragraph{Optimal step size}

The \ac{F-FNO} includes a step size parameter which specifies how many seconds
of simulation time one application of the operator will advance. A large step
size sees the model predicting far into the future possibly leading to large
errors, while a small step size means small errors have many more steps to
compound. We thus try different step sizes in \cref{fig:corr_stepsize}.

In numerical solvers, there is a close relationship between the step size in
the time dimension and the spatial resolution. Specifically, the \ac{CFL}
condition provides an optimal step size given a space discretization: $\Delta t
= C_{\text{max}} \Delta x / \lVert\mathbf{u}\rVert_{\text{max}}$. This means that if we
double the grid size, the solver should take a step that is twice as small (we
follow this approach to obtain \ac{DNS}'s Pareto frontier in
\cref{fig:pareto}). Furthermore, having step sizes smaller than what is
specified by the \ac{CFL} condition would not provide any further benefit
unless we also reduce the distance between two grid points (see purple line in
\cref{fig:corr_stepsize}). On the other hand, a step size that is too big
(e.g., bigger than 0.05 on a 64x64 grid)
will lead to stability issues in the
numerical solver.

For the \ac{F-FNO}, we find that we can take a step size that is at least an
order of magnitude bigger than the stable step size for a numerical solver.
This is the key contribution to the efficiency of neural methods. Furthermore,
there is a sweet spot for the step size -- around 0.2 on \TorusKochkov\ -- and
unlike its numerical counterpart, we find that there is no need to reduce the
step size as we train the \ac{F-FNO} on a higher spatial resolution.

\begin{figure}[t]
  \centering

\begin{subfigure}[t]{0.48\textwidth}
  \centering
  \includegraphics[width=\textwidth]{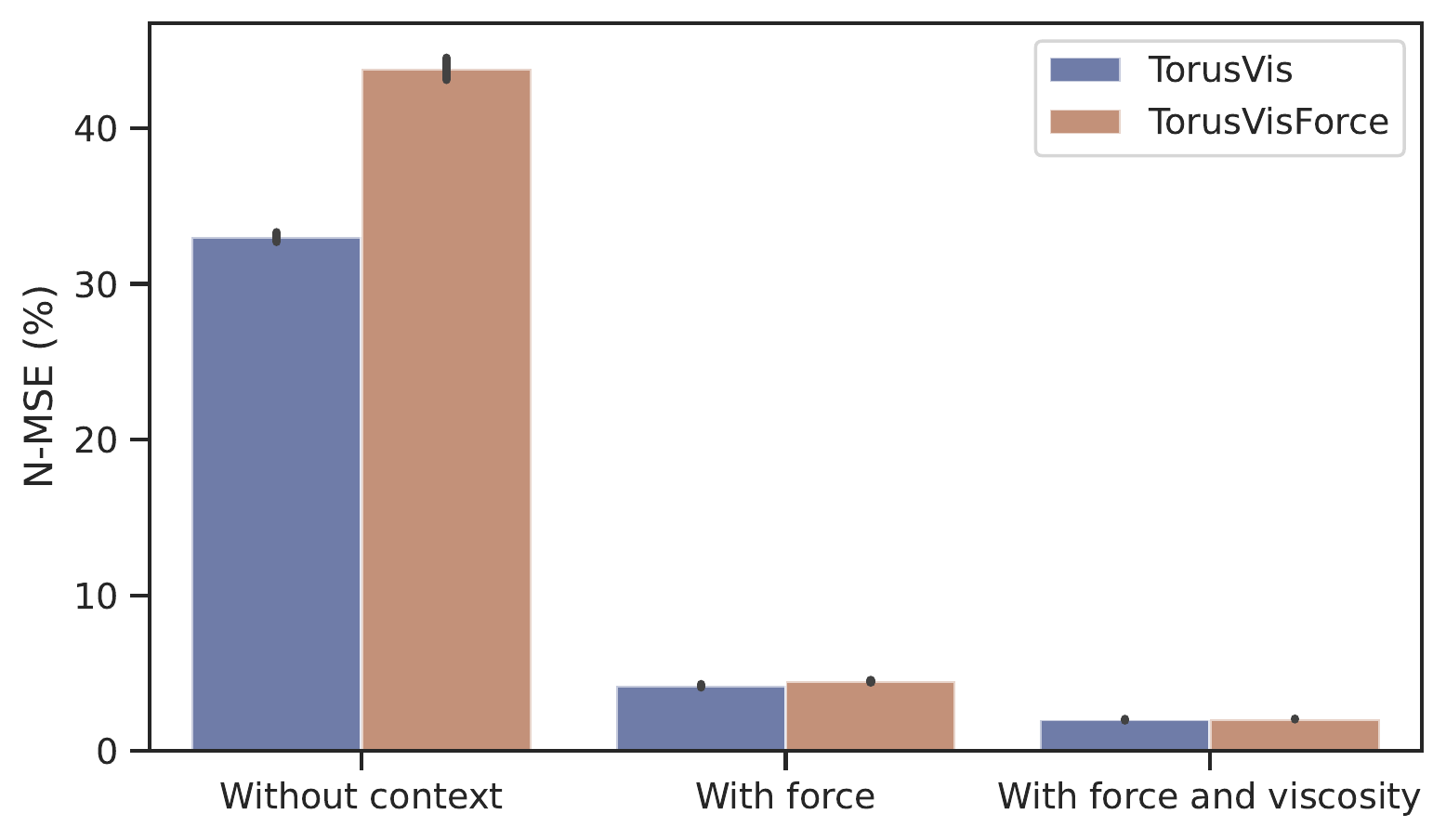}
  \caption{Performance of \ac{F-FNO} on different input features:
          having only vorticity as an input
          with no further context (first group); having vorticity
          and the force field as inputs (second group); and having
          vorticity, the force field, and viscosity as inputs
          (third group).
          The error bars are the standard deviation from
          three trials.}
  \label{fig:vis-force-ablation}
\end{subfigure}
\hfill
\begin{subfigure}[t]{0.48\textwidth}
  \centering
  \includegraphics[width=\textwidth]{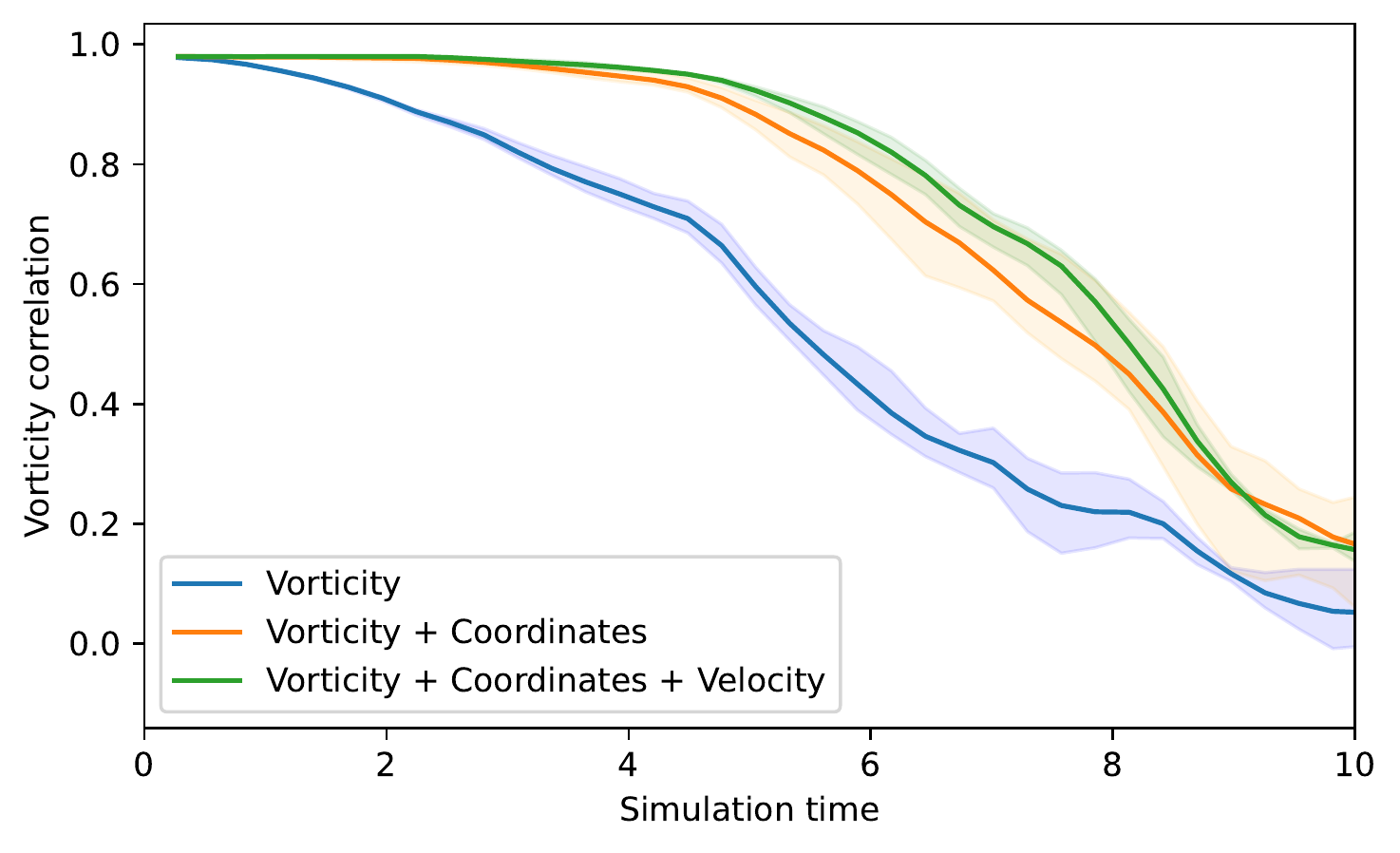}
  \caption{Effect of having the coordinates and velocity as additional
            input channels
            on \TorusKochkov. A higher line corresponds to a model that
            can correlate with the ground-truth vorticity for longer.
            Error bands correspond to min and max values from three trials.}
  \label{fig:kochkov-ablation}
\end{subfigure}
\caption{Performance of F-FNO on different contexts and input representations.}
\label{fig:flexible-inputs}
\end{figure}

\paragraph{Flexible input representations}

The \ac{F-FNO} can be trained to handle Navier-Stokes equations with
viscosities (in \TorusV) and time-varying forcing functions (in \TorusVF)
provided at inference time.
%\cref{fig:context} shows an example where our model takes these two
% contexts as additional input channels.
Our model, when given both the force and viscosity, in addition to the vorticity,
is able to achieve an error of 2\% (\cref{fig:vis-force-ablation}). If we
remove the viscosity information, the error doubles.
Removing the
forcing function from the input further increases the error by an order of magnitude.
This shows that the force has a substantial impact on the future vorticity
field, and that the \ac{F-FNO} can use information about the
forcing function to make accurate predictions.
More generally, different datasets benefit from having different input
features -- \cref{tab:reps} shows the minimum set of features to reach optimal
performance on each of them. We also find that having redundant features does
not significantly hurt the model, so there is no need to do aggressive feature
pruning in practice.

Our experiments with different input representations also reveal an interesting
performance gain from the \textit{double encoding of information}
(\cref{fig:kochkov-ablation}).
% For example, even though there is a one-to-one map between the vorticity and
% velocity fields, have both as inputs improves the \ac{F-FNO}'s performance on
% \TorusKochkov (\cref{fig:kochkov-ablation}).
All datasets benefit from the coordinate encoding -- i.e., having the $(x,y)$
coordinates as two additional input channels -- even if the positional
information is already contained in the absolute position of grid points
(indices of the input array). We hypothesize that these two positional
representations are used by different parts of the \ac{F-FNO}. The Fourier
transform uses the absolute position of the grid points and thus the Fourier
layer should have no need for the $(x,y)$ positional features. However, the
feedforward layer in the physical space is a pointwise operator and thus needs
to rely on the raw coordinate values, since it would otherwise be independent
of the absolute position of grid points.
% Finally, we find that introducing the sinusoidal position encoding,
% commonly used in language models, does not noticeably change the performance.

% There are other features which we find to have no effect on the performance
% on any dataset.
% We also try predicting the change in vorticity, inspired by the literature on
% time series prediction~\cite{Tran2021Radflow}, but this leads to worse
% performance.
% Adding the
% velocity field as additional input channels makes no difference to the
% results -- this makes sense since there is a one-to-one map between the velocity
% and vorticity field, so knowing the velocity provides no new information.

\section{Results for PDEs on point clouds and meshes}
\label{sec:resuls-mesh}

\begin{table}[t]
  \caption{Performance (N-MSE, expressed as percentage, where lower is better)
  on point clouds (\Elasticity) and structured meshes (\Airfoil and
  \Plasticity) between our \ac{F-FNO} and the previous state-of-the-art geo-FNO
  \citep{Li2022FourierNO}. Cells with a dash correspond to models which do not
  converge. The N-MSE is accompanied by the standard deviation from three
  trials. More detailed results are shown in
  \cref{tab:results_airfoil,tab:results_elasticity,tab:results_plasticity}.}

  \label{tab:meshes}
  \begin{center}
    \small{
  \begin{tabularx}{\columnwidth}{Xrrrrrr}
    \toprule
       \multirow{2}{*}{No. of layers}
     & \multicolumn{2}{c}{\Elasticity}
     & \multicolumn{2}{c}{\Airfoil}
     & \multicolumn{2}{c}{\Plasticity} \\
     & geo-FNO & F-FNO & geo-FNO & F-FNO & geo-FNO & F-FNO \\
    \midrule
    4 layer & $2.5 \pm 0.1$ & $3.16 \pm 1.29$ & $1.9 \pm 0.4$ & $0.79 \pm 0.02$ & $0.74 \pm 0.01$ & $0.48 \pm 0.02$ \\
    8 layer & $3.3 \pm 1.3$ & $2.05 \pm 0.01$ & $1.4 \pm 0.5$ & $0.64 \pm 0.01$ & $0.57 \pm 0.04$ & $0.32 \pm 0.01$ \\
    12 layer & $16.8 \pm 0.7$ & $1.96 \pm 0.02$ & $4.1 \pm 4.4$ & $0.62 \pm 0.03$ & $0.45 \pm 0.03$ & $0.25 \pm 0.01$ \\
    16 layer & $16.3 \pm 0.4$ & $1.86 \pm 0.02$ & - & $0.61 \pm 0.01$ & - & $0.22 \pm 0.00$ \\
    20 layer & $16.0 \pm 0.7$ & $1.84 \pm 0.02$ & - & $0.57 \pm 0.01$ &  -  & $0.20 \pm 0.02$ \\
    24 layer & $15.9 \pm 0.5$ & $1.74 \pm 0.03$ & - & $0.58 \pm 0.04$ & - & $0.18 \pm 0.00$ \\
    \bottomrule
  \end{tabularx}}

\end{center}
\end{table}

%\paragraph{Comparison against geo-FNO}
As shown in \cref{tab:meshes}, the geo-FNO \citep{Li2022FourierNO}, similar to
the original FNO, also suffers from network scaling. It appears to be stuck in
a local minimum beyond 8 layers in the \Elasticity problem and it completely
fails to converge beyond 12 layers in \Airfoil and \Plasticity. \Plasticity is
the only task in which the geo-FNO gets better as we go from 4 to 12 layers
($0.74\% \rightarrow 0.45\%$). In addition to the poor scaling with network
depth, we also find during our experiments that the geo-FNO can perform worse
as we increase the hidden size $H$. This indicates that there might not be
enough regularization in the model as we increase the model complexity.

Our \ac{F-FNO}, on the other hand, continues to gain performance with deeper
networks and bigger hidden size, reducing the prediction error by 31\% on the
\Elasticity point clouds ($2.51\% \rightarrow 1.74\%$) and by 57\% on the 2D
transonic flow over airfoil problem ($1.35\% \rightarrow 0.58\%$). Our Fourier
factorization particularly shines in the plastic forging problem, in which the
neural operator needs to output a 3D array, i.e., the displacement of each
point on a 2D mesh over 20 time steps. As shown in
\cref{tab:results_plasticity}, our 24-layer F-FNO with 11M parameters
outperforms the 12-layer geo-FNO with 57M parameters by 60\% ($0.45\%
\rightarrow 0.18\%$).

% !TEX root = ../main.tex

\section{Conclusion}
\label{sec:conclusion}

The Fourier transform is a powerful tool to learn neural operators that can
handle long-range spatial dependencies. By factorizing the transform, using
better residual connections, and improving the training setup, our proposed
\ac{F-FNO} outperforms the state of the art on PDEs on a variety of geometries
and domains. For future work, we are interested in examining equilibrium
properties of generalized Fourier operators with an infinite number of layers
and checking if the universal approximation property~\citep{Kovachki2021OnUA}
still holds under Fourier factorization.

\newpage

\bibliography{main}

\begin{thebibliography}{30}
\providecommand{\natexlab}[1]{#1}
\providecommand{\url}[1]{\texttt{#1}}
\expandafter\ifx\csname urlstyle\endcsname\relax
  \providecommand{\doi}[1]{doi: #1}\else
  \providecommand{\doi}{doi: \begingroup \urlstyle{rm}\Url}\fi

\bibitem[Alfonsi(2009)]{Alfonsi2009ReynoldsAveragedNE}
Giancarlo Alfonsi.
\newblock Reynolds-averaged navier–stokes equations for turbulence modeling.
\newblock \emph{Applied Mechanics Reviews}, 62:\penalty0 040802, 2009.

\bibitem[Bar-Sinai et~al.(2019)Bar-Sinai, Hoyer, Hickey, and
  Brenner]{BarSinai2019Learning}
Yohai Bar-Sinai, Stephan Hoyer, Jason Hickey, and Michael~P. Brenner.
\newblock Learning data-driven discretizations for partial differential
  equations.
\newblock \emph{Proceedings of the National Academy of Sciences}, 116\penalty0
  (31):\penalty0 15344--15349, 2019.
\newblock ISSN 0027-8424.
\newblock \doi{10.1073/pnas.1814058116}.
\newblock URL \url{https://www.pnas.org/content/116/31/15344}.

\bibitem[Bhattacharya et~al.(2020)Bhattacharya, Hosseini, Kovachki, and
  Stuart]{bhattacharya2020model}
Kaushik Bhattacharya, Bamdad Hosseini, Nikola~B Kovachki, and Andrew~M Stuart.
\newblock Model reduction and neural networks for parametric {PDE}s.
\newblock \emph{arXiv preprint arXiv:2005.03180}, 2020.

\bibitem[Chandler \& Kerswell(2013)Chandler and
  Kerswell]{Chandler2013Invariant}
Gary~J. Chandler and Rich~R. Kerswell.
\newblock Invariant recurrent solutions embedded in a turbulent two-dimensional
  kolmogorov flow.
\newblock \emph{Journal of Fluid Mechanics}, 722:\penalty0 554--595, 2013.
\newblock \doi{10.1017/jfm.2013.122}.

\bibitem[Hosseini et~al.(2016)Hosseini, Vinuesa, Schlatter, Hanifi, and
  Henningson]{Hosseini2016Direct}
S.M. Hosseini, R.~Vinuesa, P.~Schlatter, A.~Hanifi, and D.S. Henningson.
\newblock Direct numerical simulation of the flow around a wing section at
  moderate reynolds number.
\newblock \emph{International Journal of Heat and Fluid Flow}, 61:\penalty0
  117--128, 2016.
\newblock ISSN 0142-727X.
\newblock \doi{https://doi.org/10.1016/j.ijheatfluidflow.2016.02.001}.
\newblock URL
  \url{https://www.sciencedirect.com/science/article/pii/S0142727X16300169}.
\newblock SI\:TSFP9 special issue.

\bibitem[Kim et~al.(2019)Kim, Azevedo, Thuerey, Kim, Gross, and
  Solenthaler]{kim2019deep}
Byungsoo Kim, Vinicius~C Azevedo, Nils Thuerey, Theodore Kim, Markus Gross, and
  Barbara Solenthaler.
\newblock Deep fluids: A generative network for parameterized fluid
  simulations.
\newblock In \emph{Computer Graphics Forum}, volume~38, pp.\  59--70. Wiley
  Online Library, 2019.

\bibitem[Kingma \& Ba(2015)Kingma and Ba]{Kingma2015Adam}
Diederik~P. Kingma and Jimmy Ba.
\newblock Adam: A method for stochastic optimization.
\newblock In \emph{International Conference on Learning Representations}, 2015.

\bibitem[Kochkov et~al.(2021)Kochkov, Smith, Alieva, Wang, Brenner, and
  Hoyer]{Kochkove2021Machine}
Dmitrii Kochkov, Jamie~A. Smith, Ayya Alieva, Qing Wang, Michael~P. Brenner,
  and Stephan Hoyer.
\newblock Machine learning{\textendash}accelerated computational fluid
  dynamics.
\newblock \emph{Proceedings of the National Academy of Sciences}, 118\penalty0
  (21), 2021.
\newblock ISSN 0027-8424.
\newblock \doi{10.1073/pnas.2101784118}.
\newblock URL \url{https://www.pnas.org/content/118/21/e2101784118}.

\bibitem[Kovachki et~al.(2021{\natexlab{a}})Kovachki, Lanthaler, and
  Mishra]{Kovachki2021OnUA}
Nikola Kovachki, Samuel Lanthaler, and Siddhartha Mishra.
\newblock On universal approximation and error bounds for fourier neural
  operators.
\newblock \emph{Journal of Machine Learning Research}, 22\penalty0
  (290):\penalty0 1--76, 2021{\natexlab{a}}.
\newblock URL \url{http://jmlr.org/papers/v22/21-0806.html}.

\bibitem[Kovachki et~al.(2021{\natexlab{b}})Kovachki, Li, Liu, Azizzadenesheli,
  Bhattacharya, Stuart, and Anandkumar]{Kovachki2021NeuralOL}
Nikola~B. Kovachki, Zong-Yi Li, Burigede Liu, Kamyar Azizzadenesheli, Kaushik
  Bhattacharya, Andrew Stuart, and Animashree Anandkumar.
\newblock Neural operator: Learning maps between function spaces.
\newblock \emph{ArXiv}, abs/2108.08481, 2021{\natexlab{b}}.

\bibitem[Lee-Thorp et~al.(2021)Lee-Thorp, Ainslie, Eckstein, and
  Onta{\~n}{\'o}n]{LeeThorp2021FNetMT}
J.~Lee-Thorp, Joshua Ainslie, Ilya Eckstein, and Santiago Onta{\~n}{\'o}n.
\newblock Fnet: Mixing tokens with fourier transforms.
\newblock \emph{ArXiv}, abs/2105.03824, 2021.

\bibitem[Lesieur \& M{\'e}tais(1996)Lesieur and M{\'e}tais]{Lesieur1996NewTI}
Marcel~R. Lesieur and Olivier M{\'e}tais.
\newblock New trends in large-eddy simulations of turbulence.
\newblock \emph{Annual Review of Fluid Mechanics}, 28:\penalty0 45--82, 1996.

\bibitem[Li et~al.(2022)Li, Huang, Liu, and Anandkumar]{Li2022FourierNO}
Zong-Yi Li, Daniel~Z. Huang, Burigede Liu, and Anima Anandkumar.
\newblock Fourier neural operator with learned deformations for {PDE}s on
  general geometries.
\newblock \emph{ArXiv}, abs/2207.05209, 2022.

\bibitem[Li et~al.(2020{\natexlab{a}})Li, Kovachki, Azizzadenesheli, Liu,
  Bhattacharya, Stuart, and Anandkumar]{Li2020MultipoleGN}
Zongyi Li, Nikola~B. Kovachki, K.~Azizzadenesheli, Burigede Liu,
  K.~Bhattacharya, Andrew Stuart, and Anima Anandkumar.
\newblock Multipole graph neural operator for parametric partial differential
  equations.
\newblock \emph{ArXiv}, abs/2006.09535, 2020{\natexlab{a}}.

\bibitem[Li et~al.(2020{\natexlab{b}})Li, Kovachki, Azizzadenesheli, Liu,
  Bhattacharya, Stuart, and Anandkumar]{Li2020NeuralOG}
Zongyi Li, Nikola~B. Kovachki, K.~Azizzadenesheli, Burigede Liu,
  K.~Bhattacharya, Andrew Stuart, and Anima Anandkumar.
\newblock Neural operator: Graph kernel network for partial differential
  equations.
\newblock \emph{ArXiv}, abs/2003.03485, 2020{\natexlab{b}}.

\bibitem[Li et~al.(2021{\natexlab{a}})Li, Kovachki, Azizzadenesheli, Liu,
  Bhattacharya, Stuart, and Anandkumar]{Li2021Fourier}
Zongyi Li, Nikola~B. Kovachki, K.~Azizzadenesheli, Burigede Liu,
  K.~Bhattacharya, Andrew Stuart, and Anima Anandkumar.
\newblock Fourier neural operator for parametric partial differential
  equations.
\newblock In \emph{International Conference on Learning Representations},
  2021{\natexlab{a}}.
\newblock URL \url{https://openreview.net/forum?id=c8P9NQVtmnO}.

\bibitem[Li et~al.(2021{\natexlab{b}})Li, Kovachki, Azizzadenesheli, Liu,
  Bhattacharya, Stuart, and Anandkumar]{Li2021MarkovNO}
Zongyi Li, Nikola~B. Kovachki, K.~Azizzadenesheli, Burigede Liu,
  K.~Bhattacharya, Andrew Stuart, and Anima Anandkumar.
\newblock Markov neural operators for learning chaotic systems.
\newblock \emph{ArXiv}, abs/2106.06898, 2021{\natexlab{b}}.

\bibitem[Loshchilov \& Hutter(2017)Loshchilov and Hutter]{Loshchilov2017SGDR}
Ilya Loshchilov and Frank Hutter.
\newblock {SGDR:} stochastic gradient descent with warm restarts.
\newblock In \emph{5th International Conference on Learning Representations,
  {ICLR} 2017, Toulon, France, April 24-26, 2017, Conference Track
  Proceedings}. OpenReview.net, 2017.
\newblock URL \url{https://openreview.net/forum?id=Skq89Scxx}.

\bibitem[Loshchilov \& Hutter(2019)Loshchilov and
  Hutter]{Loshchilov2018Decoupled}
Ilya Loshchilov and Frank Hutter.
\newblock Decoupled weight decay regularization.
\newblock In \emph{International Conference on Learning Representations}, 2019.

\bibitem[Lu et~al.(2022)Lu, Meng, Cai, Mao, Goswami, Zhang, and
  Karniadakis]{Lu2022ACA}
Lu~Lu, Xuhui Meng, Shengze Cai, Zhiping Mao, Somdatta Goswami, Zhongqiang
  Zhang, and George~Em Karniadakis.
\newblock A comprehensive and fair comparison of two neural operators (with
  practical extensions) based on fair data.
\newblock \emph{Computer Methods in Applied Mechanics and Engineering}, 2022.

\bibitem[Obiols-Sales et~al.(2020)Obiols-Sales, Vishnu, Malaya, and
  Chandramowliswharan]{obiols2020cfdnet}
Octavi Obiols-Sales, Abhinav Vishnu, Nicholas Malaya, and Aparna
  Chandramowliswharan.
\newblock Cfdnet: A deep learning-based accelerator for fluid simulations.
\newblock In \emph{Proceedings of the 34th ACM International Conference on
  Supercomputing}, pp.\  1--12, 2020.

\bibitem[Pascon(2019)]{Pascon2019LargeDA}
Jo{\~a}o~Paulo Pascon.
\newblock Large deformation analysis of plane-stress hyperelastic problems via
  triangular membrane finite elements.
\newblock \emph{International Journal of Advanced Structural Engineering},
  2019.

\bibitem[Paszke et~al.(2017)Paszke, Gross, Chintala, Chanan, Yang, DeVito, Lin,
  Desmaison, Antiga, and Lerer]{Paszke2017Automatic}
Adam Paszke, Sam Gross, Soumith Chintala, Gregory Chanan, Edward Yang, Zachary
  DeVito, Zeming Lin, Alban Desmaison, Luca Antiga, and Adam Lerer.
\newblock Automatic differentiation in {PyTorch}.
\newblock In \emph{NIPS Autodiff Workshop}, 2017.

\bibitem[Poli et~al.(2022)Poli, Massaroli, Berto, Park, Dao, Re, and
  Ermon]{poli2022transform}
Michael Poli, Stefano Massaroli, Federico Berto, Jinkyoo Park, Tri Dao,
  Christopher Re, and Stefano Ermon.
\newblock Transform once: Efficient operator learning in frequency domain.
\newblock In \emph{ICML 2022 2nd AI for Science Workshop}, 2022.
\newblock URL \url{https://openreview.net/forum?id=x1fNT5yj41N}.

\bibitem[Sanchez-Gonzalez et~al.(2020)Sanchez-Gonzalez, Godwin, Pfaff, Ying,
  Leskovec, and Battaglia]{SanchezGonzalez2020LearningTS}
Alvaro Sanchez-Gonzalez, Jonathan Godwin, Tobias Pfaff, Rex Ying, Jure
  Leskovec, and Peter~W. Battaglia.
\newblock Learning to simulate complex physics with graph networks.
\newblock In \emph{Proceedings of the 37th International Conference on Machine
  Learning}, ICML'20. JMLR.org, 2020.

\bibitem[Tompson et~al.(2017)Tompson, Schlachter, Sprechmann, and
  Perlin]{tompson2017accelerating}
Jonathan Tompson, Kristofer Schlachter, Pablo Sprechmann, and Ken Perlin.
\newblock Accelerating eulerian fluid simulation with convolutional networks.
\newblock In \emph{International Conference on Machine Learning}, pp.\
  3424--3433. PMLR, 2017.

\bibitem[Um et~al.(2020)Um, Brand, Fei, Holl, and Thuerey]{Um2020Solver}
Kiwon Um, Robert Brand, Yun Fei, Philipp Holl, and Nils Thuerey.
\newblock {Solver-in-the-Loop: Learning from Differentiable Physics to Interact
  with Iterative {PDE}-Solvers}.
\newblock \emph{Advances in Neural Information Processing Systems}, 2020.

\bibitem[Vaswani et~al.(2017)Vaswani, Shazeer, Parmar, Uszkoreit, Jones, Gomez,
  Kaiser, and Polosukhin]{Vaswani2017Attention}
Ashish Vaswani, Noam Shazeer, Niki Parmar, Jakob Uszkoreit, Llion Jones,
  Aidan~N Gomez, \L~ukasz Kaiser, and Illia Polosukhin.
\newblock Attention is all you need.
\newblock In I.~Guyon, U.~Von Luxburg, S.~Bengio, H.~Wallach, R.~Fergus,
  S.~Vishwanathan, and R.~Garnett (eds.), \emph{Advances in Neural Information
  Processing Systems}, volume~30. Curran Associates, Inc., 2017.
\newblock URL
  \url{https://proceedings.neurips.cc/paper/2017/file/3f5ee243547dee91fbd053c1c4a845aa-Paper.pdf}.

\bibitem[Wandel et~al.(2020)Wandel, Weinmann, and Klein]{wandel2020learning}
Nils Wandel, Michael Weinmann, and Reinhard Klein.
\newblock Learning incompressible fluid dynamics from scratch--towards fast,
  differentiable fluid models that generalize.
\newblock \emph{arXiv preprint arXiv:2006.08762}, 2020.

\bibitem[Wang et~al.(2020)Wang, Kashinath, Mustafa, Albert, and
  Yu]{wang2020towards}
Rui Wang, Karthik Kashinath, Mustafa Mustafa, Adrian Albert, and Rose Yu.
\newblock Towards physics-informed deep learning for turbulent flow prediction.
\newblock In \emph{Proceedings of the 26th ACM SIGKDD International Conference
  on Knowledge Discovery \& Data Mining}, pp.\  1457--1466, 2020.

\end{thebibliography}
\bibliographystyle{iclr2023_conference}

\newpage
% !TEX root = ../main.tex

\counterwithin{figure}{section}
\counterwithin{table}{section}
\appendix

% \renewcommand{\thefigure}{A\arabic{figure}}
% \renewcommand{\thetable}{A\arabic{table}}
% \setcounter{figure}{0}
% \setcounter{table}{0}

% \section{Code and Datasets}
% \label{ssec:code}

% The code, pretrained models, and datasets are licensed under the MIT License. They
% can be found at \url{https://anonymized}.

\section{Appendix}
% \label{ssec:further_results}

% \begin{figure}[h!]
%   \centering
%   \includegraphics[width=0.4\linewidth]{figures/context}
%     \caption{\ac{F-FNO} solves the Navier-Stokes equations
%                with a range of parameters without needing to retrain, by
%                taking viscosity $\nu$ and forcing function $f_t$ as input.}
%       %On the more general setting, the \ac{F-FNO} can take
%       %additional contexts such as the viscosity and the time-depdendent
%       %forcing function as input.}
%       \label{fig:context}
% \end{figure}

% Detailed results, including the normalized mean squared error, along with complexity
% measures such as the parameter count and inference time, are shown in
% \cref{tab:results_torus_const}.

% \subsection{Datasets}
% \label{apdx:dataset}

% !TEX root = ../main.tex

\begin{table*}[h]
    \caption{An overview of the four fluid dynamics datasets on regular grids.
             Our newly generated datasets, \TorusV and
             \TorusVF, contain simulation data with a more variety of viscosities
             and forces than \TorusZongyi \citep{Li2021Fourier} and \TorusKochkov
             \citep{Kochkove2021Machine}.
             Note that \citet{Li2021Fourier}
             did not generate a validation set.}

    \label{tab:datasets_torus}
    \begin{center}\small{
    \begin{tabularx}{\textwidth}{Xrrrlcc}
      \toprule
      \multirow{2}{*}{Dataset}
        & \multirow{2}{*}{\makecell[r]{ Train / valid / test \\ split }}
        & \multirow{2}{*}{\makecell[r]{Trajectory \\ length}}
        & \multirow{2}{*}{Domain}
        & \multirow{2}{*}{Viscosity}
        & \multicolumn{2}{c}{Force varying across} \\
      \cmidrule(r){6-7}
      & & & & & samples     & time \\
      \midrule
      \TorusZongyi
        & 1000 / 0 / 200
        & 20
        & $[0, 1]$
        & $ \nu = 10^{-5} $
        & &  \\
      \TorusKochkov 
        & 32 / 4 / 4
        & 34
        & $[0, 2\pi]$
        & $\nu = 10^{-3}$
        &  &  \\
      \TorusV
        & 1000 / 200 / 200
        & 20
        & $[0, 1]$
        & $ \nu \in [10^{-5}, 10^{-4}) $
        & \cmark & \\
      \TorusVF
        & 1000 / 200 / 200
        & 20
        & $[0, 1]$
        & $ \nu \in [10^{-5}, 10^{-4}) $
        & \cmark & \cmark \\
      \bottomrule
    \end{tabularx}}
  \end{center}
  \end{table*}

\begin{table*}[h]
    \caption{An overview of the three PDE datasets on irregular geometries. These
             datasets were generated by \citet{Li2022FourierNO}.}

    \label{tab:datasets_mesh}
    \begin{center}\small{
    \begin{tabularx}{\textwidth}{Xrrrll}
      \toprule
      Dataset & Train & Valid & Test & Governing equation & Problem dimension \\
      \midrule
      \Elasticity
        & 1000
        & 200
        & 200 
        & Equation of a solid body
        & point cloud on 2D unit cell \\
      \Airfoil
        & 1000
        & 200
        & 200
        & Euler's equation 
        & 2D structured mesh \\
      \Plasticity
        & 827
        & 80
        & 80 
        & Equation of a solid body 
        & 2D structured mesh over 1D time \\
      \bottomrule
    \end{tabularx}}
  \end{center}
  \end{table*}

\begin{figure*}[h]
  \centering
  \includegraphics[width=\linewidth]{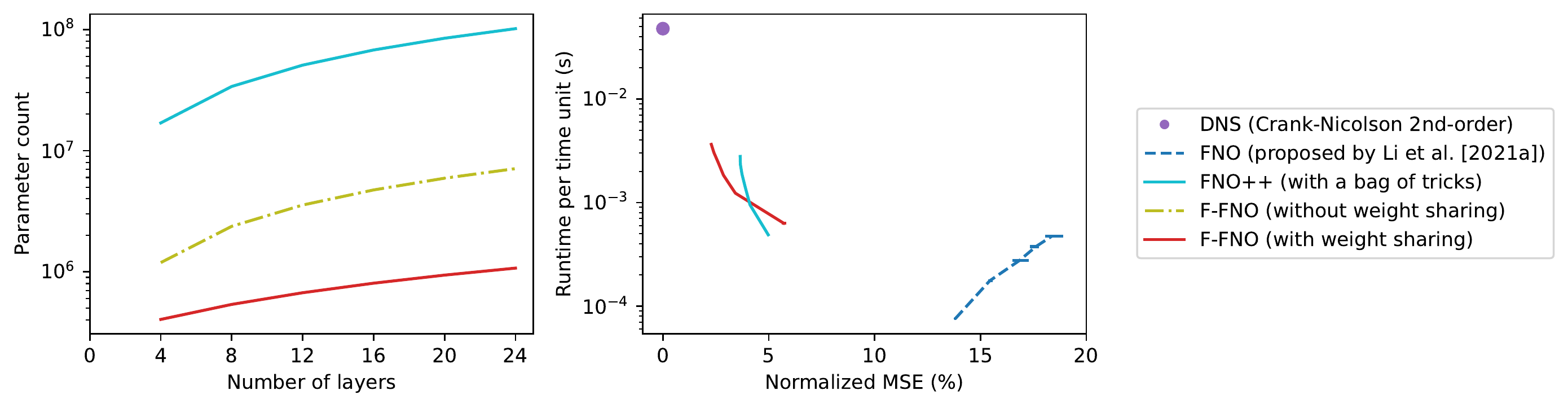}
  \
  \captionsetup[subfigure]{position=bottom,skip=2pt}
  \parbox{0.4\linewidth}{\subcaption{} \label{fig:param_count}}
  \parbox{0.3\linewidth}{\subcaption{} \label{fig:inference_time}}\hfill
  \
  \caption{The resource usage of four model variants, in terms of
  (a) the parameter count and
  (b) inference time (the time it takes to run one second of simulation).
  Error bars, when applicable, show
  the min and max values over three trials. In (b), as we move along a line,
  we increase the number of layers. We observe that only our model variants (F-FNO)
  have the desired slope, that is, as we use more resources (increasing the inference time),
  we obtain better predictions.}
  \label{fig:complexity}
\end{figure*}

\pagebreak

\begin{figure}[h]
  \centering
  \begin{subfigure}[b]{0.49\textwidth}
    \centering
    \includegraphics[width=\textwidth]{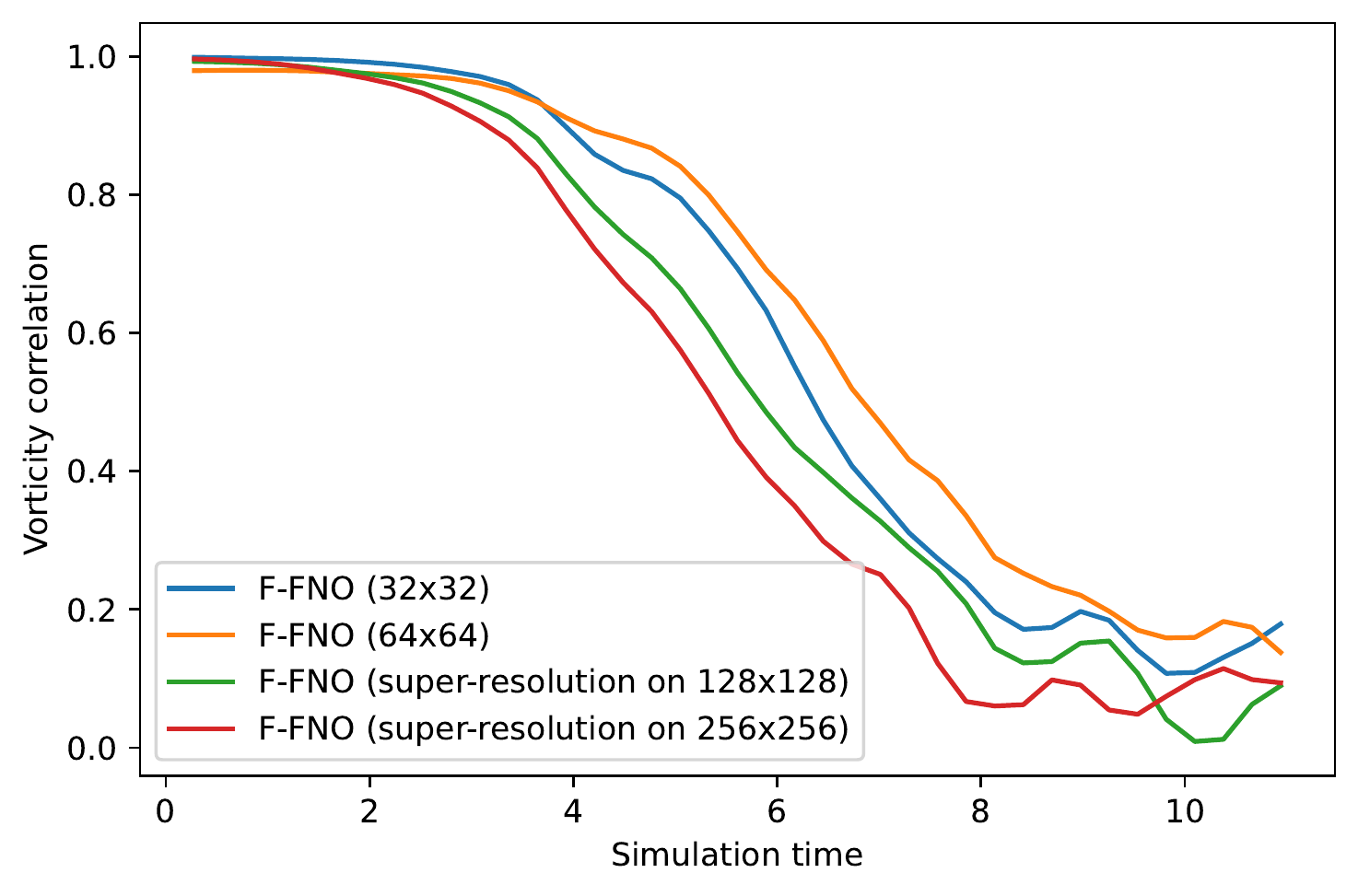}
    \caption{Zero-shot super-resolution performance of \ac{F-FNO}. We train the
    model on 32x32 and 64x64 grids of \TorusKochkov, and evaluate on the larger
    128x128 and 256x256 grids. We observe some degradation in the correlation
    with the ground truths on unseen grid sizes.}
    \label{fig:super_64}
  \end{subfigure}
  \hfill
  \begin{subfigure}[b]{0.49\textwidth}
    \centering
  \includegraphics[width=\textwidth]{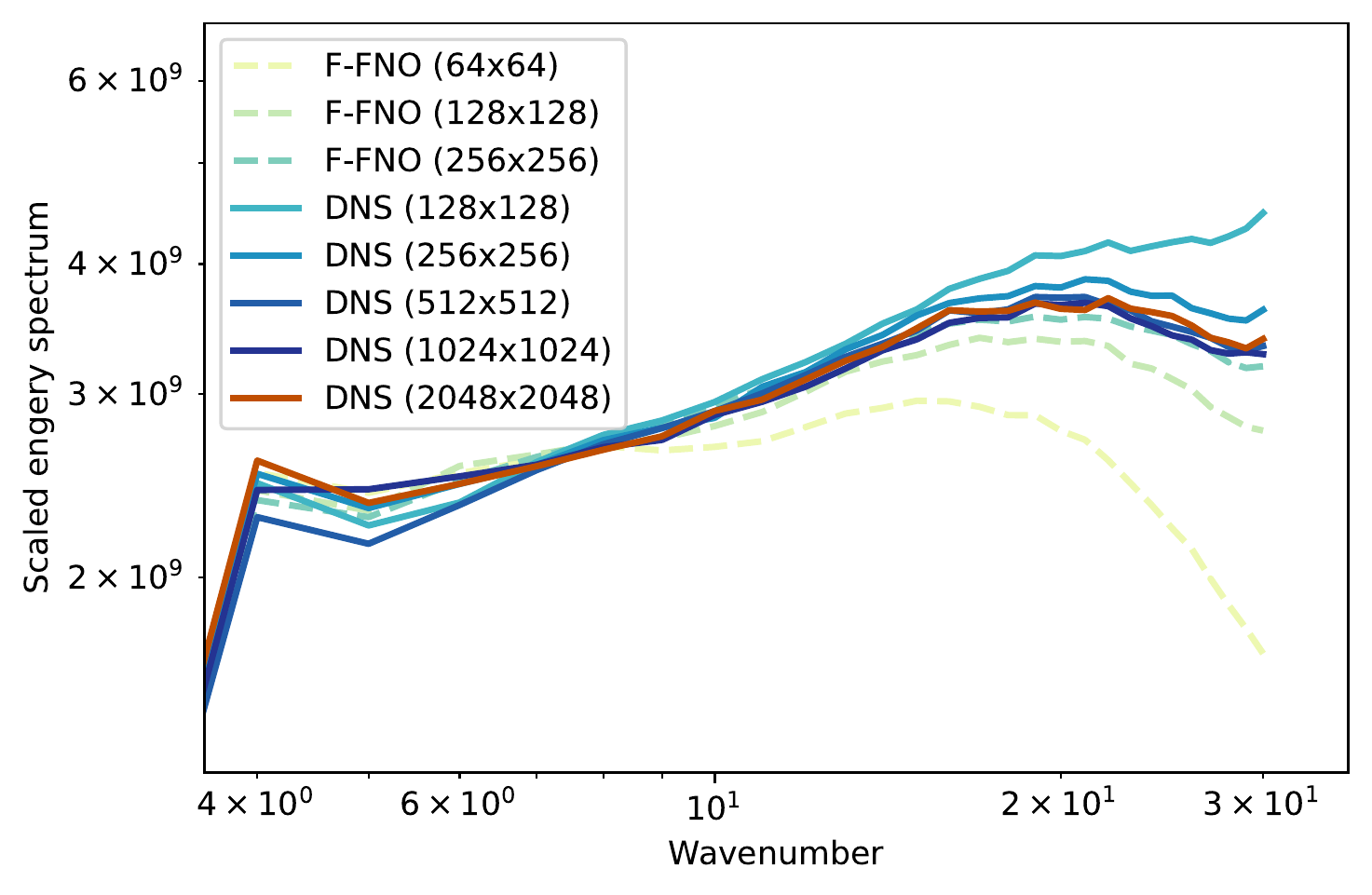}
  \caption{Energy spectra of \ac{F-FNO} and \ac{DNS} on various grid
  sizes. The spectra are computed by averaging the kinetic energy for each
  wavenumber between $t=12$ and $t=34$, when the predictions from all methods
  have decorrelated with the ground truths.}
  \label{fig:energy}
  \end{subfigure}
  \caption{Performance of F-FNO on zero-shot superresolution and its ability to
           capture the energy spectrum of DNS on \TorusKochkov.}
  \label{fig:zero_energy}
\end{figure}

\paragraph{Zero-shot super-resolution}

In \cref{fig:super_64}, we train the \ac{F-FNO} once on 32x32 and 64x64 grids
from \TorusKochkov, and then perform inference and evaluation on 128x128 and
256x256 grids. This extends the super-resolution setting presented by
\citet{Li2021Fourier} as they only worked on simple PDEs such as the 1D
Burger's equation and the 2D Darcy flow.  We find that although the \ac{F-FNO} can
do zero-shot super-resolution -- unlike a traditional CNN which by design cannot
even accept inputs of variable size -- its performance does degrade on grid
sizes not seen during training. This is seen by the lower vorticity correlation
of the super-resolution \ac{F-FNO} settings in \cref{fig:super_64}. We posit
that the super-resolution performance could be improved by training on a
variety of grid sizes (e.g., by downsampling each training
example to a random size). We leave such exploration for future work.

% We also note that after unrolling for 10 seconds, the predictions have
% completely decorrelated with the ground truths.

\paragraph{Capturing the energy spectrum}

In addition to having a high vorticity correlation, a good model should also
produce predictions with an energy spectrum similar to the most accurate
numerical methods. Given the Fourier transform of a velocity field
$\hat{\mathbf{u}} = \text{FFT}(\mathbf{u})$, we can compute, for each
wavenumber $k$, the corresponding kinetic energy as $E(k) = \frac{1}{2} \lVert
\hat{\mathbf{u}}_k \rVert^2$. \cref{fig:energy} shows the energy spectrum of
both the \ac{F-FNO} and \ac{DNS} at different resolutions. These multiple
\ac{DNS} resolutions are included both as a reference solution in the case of
\ac{DNS} 2048x2048, and to demonstrate that increasing the resolution of
\ac{DNS} further is not likely to substantially change the energy spectrum.

We observe that compared to \ac{DNS} on 2048x2048, the \ac{F-FNO} trained on
64x64 grids produces an energy spectrum that has substantially lower energy at
high wavenumbers. This is expected as at this spatial resolution we only
select the top 16 modes in each Fourier layer. Even so, the \ac{F-FNO} can
still capture the long term trend much better than running \ac{DNS} on a grid
four times its size (see \cref{fig:pareto}). As we select more Fourier modes on
bigger grids (top 32 modes on 128x128 grids and top 64 modes on 256x256 grids),
the energy spectrum produced converges towards that of the reference solution (DNS on
2048x2048). This gives some indication that the \ac{F-FNO} is able to accurately
predict both high and low frequency details.

\paragraph{Effect of using cosine transforms}

\begin{figure}[p]
  \centering
  \begin{subfigure}[b]{0.49\textwidth}
    \centering
    \includegraphics[width=\textwidth]{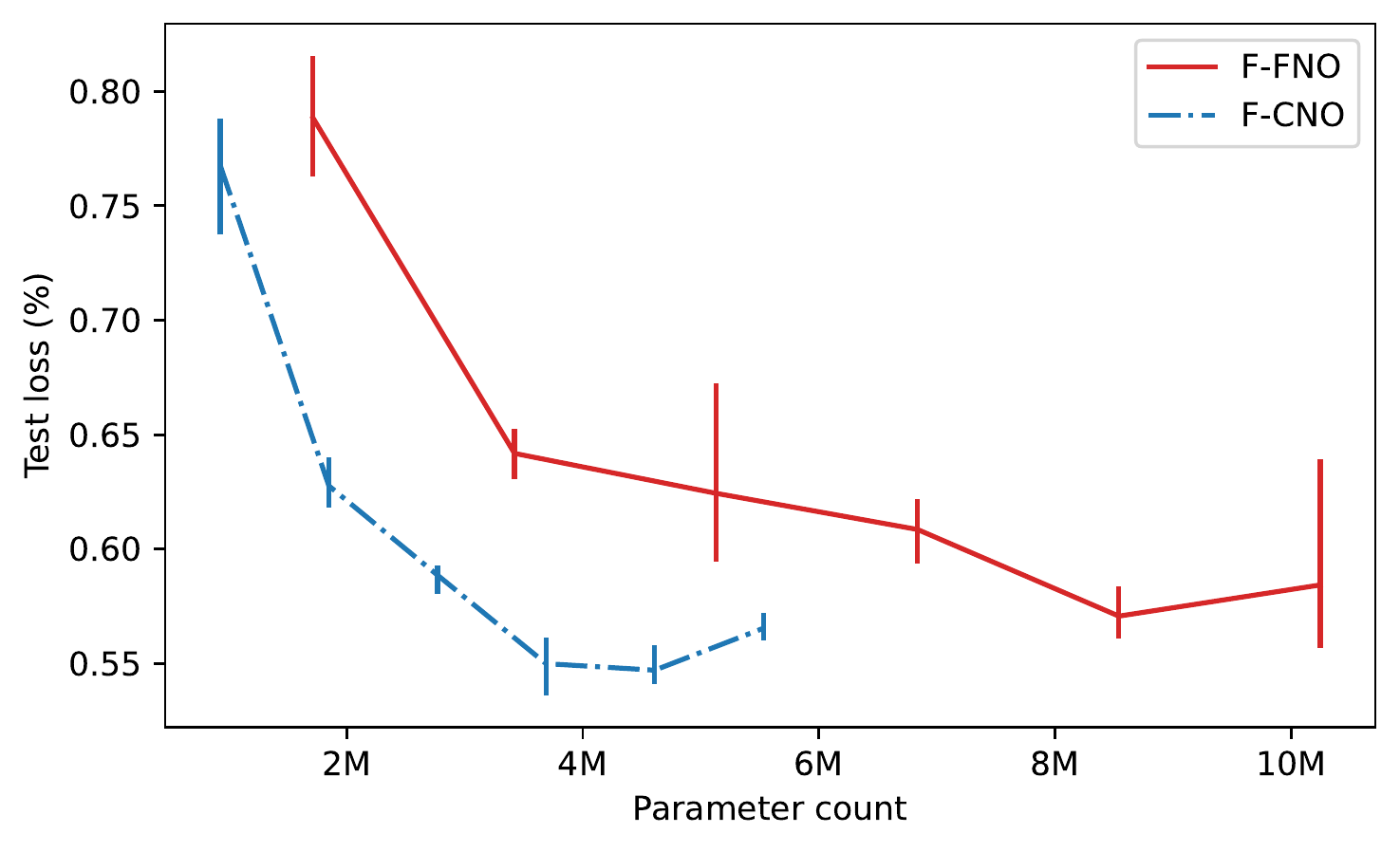}
    \caption{Airfoil}
    \label{fig:fcno_airfoil}
  \end{subfigure}
  \hfill
  \begin{subfigure}[b]{0.49\textwidth}
    \centering
  \includegraphics[width=\textwidth]{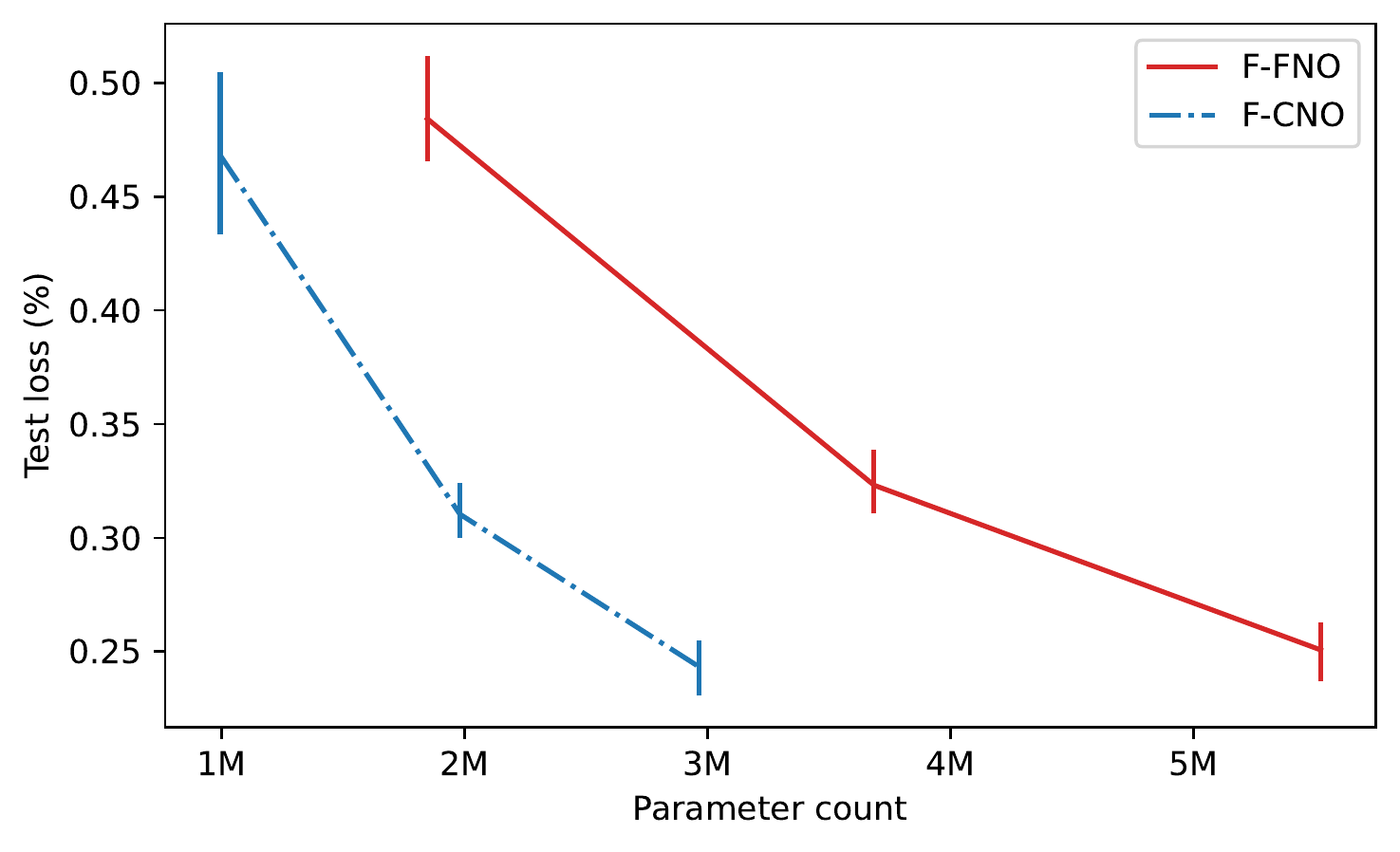}
  \caption{Plasticity}
  \label{fig:fcno_plasiticty}
  \end{subfigure}
  \caption{Effect of the cosine transform on \Airfoil and \Plasticity. We plot
  the test loss (y-axis) against the model parameter count (x-axis). Error bars
  show the min-max values from three trials. As we move a long each line, we
  make the network deeper, which increases the number of parameters. On
  \Airfoil (a), the F-CNO outperforms the F-FNO at deeper layers. On
  \Plasticity (b), the performance between the two is mostly similar for the
  same depth. Since cosine transforms are real-valued, the F-CNO requires only
  half as many parameters as the F-FNO.}
  \label{fig:fcno}
\end{figure}

As an alternative to the Fourier transform, \citet{poli2022transform} proposed
using the cosine transform, which has the advantage of being real-valued, thus
halving the number of parameters. Let the Factorized Cosine Neural Operator
(F-CNO) be the operator where the Fourier transform is replaced with the cosine
transform. In \cref{fig:fcno}, we observe that on \Airfoil, the F-CNO
outperforms the F-FNO especially at deeper layers. On \Plasticity, the F-CNO
performs comparably to the F-FNO on the same depth, while using fewer
parameters. We have not had much success in training the F-CNO on torus
datasets such as \TorusKochkov. We leave the investigation of how stable the
cosine transform is on different domains to future work.

\begin{table}[h]
    \caption{Detailed performance on \TorusZongyi. These results are used to
    generate \cref{fig:torus_li_performance} in the main paper.
    We run three trials for each experiment, each with a different random seed.
    We report the mean N-MSE from the three trials, along with the min and max value.
    A dash indicates that the data is not available.
    % Our
    % reruns use an improved implementation from the original authors, where
    % batch norm layers are removed. We fix the width to 32, selecting the top
    % 16 modes, fix batch size to 4, and train for 50 epochs.
    }

    \label{tab:results_torus_const}
    \begin{center}
    {\small
    \begin{tabularx}{\columnwidth}{Xrrrrrr}
      \toprule
       & \multirow{2}{*}{\makecell[r]{No. of \\ layers}}
       & \multirow{2}{*}{\makecell[r]{No. of \\ parameters}}
       & \multicolumn{3}{c}{N-MSE (\%)}
       & \multirow{2}{*}{\makecell[r]{Training \\ time (h)}} \\
      \cmidrule(r){4-6}
       & & & \makecell[r]{Mean}
       & \makecell[r]{Min}
       & \makecell[r]{Max}
       \\
      \midrule
      ResNet \citep{Li2021Fourier} & - & 266,641 & 27.53 & - & - \\
      TF-Net \citep{Li2021Fourier} & - & 7,451,724 & 22.68 & - & -  \\
      U-Net \citep{Li2021Fourier} & - & 7,451,724 & 19.82 & - & -  \\
      FNO \citep{Li2021Fourier} & 4 & 414,517 & 15.56 & - & - \\
      \midrule
      \multirow{5}{*}{FNO (reproduced)}
      & 4 & 926,357 & 13.80 & 13.75 & 13.83 &  2 \\
      & 8 & 1,849,637 & 15.45 & 15.40 & 15.50 &  3 \\
      & 12 & 2,772,917 & 16.86 & 16.55 & 17.27 &  5 \\
      & 16 & 3,696,197 & 17.59 & 17.41 & 17.87 &  6 \\
      & 20 & 4,619,477 & 18.44 & 17.91 & 19.24 &  7 \\
      & 24 & \multicolumn{5}{c}{does not converge} \\
     \midrule
     \multirow{5}{*}{FNO-TF (FNO with teacher forcing)}
      & 4 & 926,357 & 12.82 & 12.50 & 13.13 &  2 \\
      & 8 & 1,849,637 & 11.05 & 10.98 & 11.10 &  3 \\
      & 12 & 2,772,917 & 11.83 & 11.74 & 11.91 &  5 \\
      & 16 & 3,696,197 & 14.10 & 13.55 & 14.80 &  6 \\
      & 20 & 4,619,477 & 17.17 & 15.40 & 19.64 &  7 \\
      & 24 & \multicolumn{5}{c}{does not converge} \\
     \midrule
     \multirow{5}{*}{FNO-M (FNO-TF with Markov assumption)}
      & 4 & 926,177 & 10.94 & 10.47 & 11.35 &  1 \\
      & 8 & 1,849,457 & 9.22 & 8.48 & 9.67 &  2 \\
      & 12 & 2,772,737 & 9.15 & 9.02 & 9.25 &  3 \\
      & 16 & 3,696,017 & 10.39 & 9.61 & 10.81 &  4 \\
      & 20 & 4,619,297 & 11.73 & 11.50 & 12.12 &  4 \\
      & 24 & \multicolumn{5}{c}{does not converge} \\
     \midrule
     \multirow{5}{*}{FNO-R (FNO-M with improved residuals)}
      & 4 & 926,177 & 10.37 & 10.08 & 10.69 &  1 \\
      & 8 & 1,849,457 & 8.36 & 8.22 & 8.46 &  2 \\
      & 12 & 2,772,737 & 7.88 & 7.78 & 7.95 &  3 \\
      & 16 & 3,696,017 & 7.90 & 7.48 & 8.25 &  4 \\
      & 20 & 4,619,297 & 7.38 & 7.34 & 7.45 &  5 \\
      & 24 & 5,542,577 & 7.33 & 7.31 & 7.36 &  5 \\
     \midrule
     \multirow{6}{*}{FNO++ (FNO-R with bags of tricks)}
      & 4 & 16,919,746 & 5.05 & 5.02 & 5.06 &  1 \\
      & 8 & 33,830,594 & 4.16 & 4.11 & 4.19 &  2 \\
      & 12 & 50,741,442 & 3.90 & 3.85 & 3.93 &  3 \\
      & 16 & 67,652,290 & 3.82 & 3.77 & 3.84 &  4 \\
      & 20 & 84,563,138 & 3.72 & 3.65 & 3.79 &  5 \\
      & 24 & 101,473,986 & 3.73 & 3.70 & 3.76 &  5 \\
     \midrule
     \multirow{6}{*}{F-FNO (FNO++ with Fourier factorization)}
      & 4 & 1,191,106 & 5.05 & 5.00 & 5.09 &  1 \\
      & 8 & 2,373,314 & 3.22 & 3.17 & 3.29 &  2 \\
      & 12 & 3,555,522 & 2.75 & 2.70 & 2.77 &  3 \\
      & 16 & 4,737,730 & 2.58 & 2.57 & 2.59 &  4 \\
      & 20 & 5,919,938 & 2.39 & 2.37 & 2.42 &  5 \\
      & 24 & 7,102,146 & 2.41 & 2.37 & 2.47 &  6 \\
     \midrule
     \multirow{6}{*}{F-FNO-WS (F-FNO with weight sharing)}
      & 4 & 404,674 & 5.74 & 5.69 & 5.79 &  1 \\
      & 8 & 538,306 & 3.43 & 3.42 & 3.44 &  2 \\
      & 12 & 671,938 & 2.87 & 2.84 & 2.90 &  3 \\
      & 16 & 805,570 & 2.56 & 2.54 & 2.57 &  4 \\
      & 20 & 939,202 & 2.42 & 2.38 & 2.45 &  5 \\
      & 24 & 1,072,834 & 2.37 & 2.31 & 2.45 &  6 \\
      \bottomrule
    \end{tabularx}
    }
  \end{center}
  \end{table}

\begin{table}[h]
    \caption{Detailed performance on \Airfoil. These results are more detailed version of
    \cref{tab:meshes} in the main paper.
    We run three trials for each experiment, each with a different random seed.
    We report the mean N-MSE from the three trials, along with the min and max value.
    % Our
    % reruns use an improved implementation from the original authors, where
    % batch norm layers are removed. We fix the width to 32, selecting the top
    % 16 modes, fix batch size to 4, and train for 50 epochs.
    }

    \label{tab:results_airfoil}
    \begin{center}
    {\small
    \begin{tabularx}{\columnwidth}{Xrrrrrr}
      \toprule
       & \multirow{2}{*}{\makecell[r]{No. of \\ layers}}
       & \multirow{2}{*}{\makecell[r]{No. of \\ parameters}}
       & \multicolumn{3}{c}{N-MSE (\%)}
       & \multirow{2}{*}{\makecell[r]{Training \\ time (h)}} \\
      \cmidrule(r){4-6}
       & & & \makecell[r]{Mean}
       & \makecell[r]{Min}
       & \makecell[r]{Max}
       \\
      \midrule
    \multirow{3}{*}{geo-FNO (reproduced)}
     & 4 & 2,368,033 & 1.87 & 1.40 & 2.27 &  4 \\
     & 8 & 4,731,553 & 1.35 & 1.02 & 2.00 &  5 \\
     & 12 & 7,095,073 & 4.11 & 0.92 & 10.29 &  4 \\
    \midrule
    \multirow{5}{*}{F-FNO}
      & 4 & 1,715,458 & 0.79 & 0.76 & 0.82 &  3 \\
      & 8 & 3,421,954 & 0.64 & 0.63 & 0.65 &  4 \\
      & 12 & 5,128,450 & 0.62 & 0.59 & 0.67 &  5 \\
      & 16 & 6,834,946 & 0.61 & 0.59 & 0.62 &  5 \\
      & 20 & 8,541,442 & 0.57 & 0.56 & 0.58 &  4 \\
      & 24 & 10,247,938 & 0.58 & 0.56 & 0.64 &  4 \\
    \midrule
    \multirow{5}{*}{F-FNO-WS (F-FNO with weight sharing)}
     & 4 & 535,810 & 0.98 & 0.90 & 1.03 &  0.4 \\
     & 8 & 669,442 & 0.72 & 0.70 & 0.75 &  0.7 \\
     & 12 & 803,074 & 0.68 & 0.66 & 0.70 &  1 \\
     & 16 & 936,706 & 0.67 & 0.63 & 0.70 &  1 \\
     & 20 & 1,070,338 & 0.64 & 0.63 & 0.66 &  2 \\
     & 24 & 1,203,970 & 0.66 & 0.60 & 0.70 &  2 \\
      \bottomrule
    \end{tabularx}
    }
  \end{center}
  \end{table}

\begin{table}[h]
    \caption{Detailed performance on \Elasticity. These results are more detailed version of
    \cref{tab:meshes} in the main paper.
    We run three trials for each experiment, each with a different random seed.
    We report the mean N-MSE from the three trials, along with the min and max value.
    Note that for a given layer, our F-FNO (whether with weight sharing or without)
    has slightly more parameters than the geo-FNO. This is due to the F-FNO using
    a bigger hidden size $H$. We find that on the geo-FNO, increasing its hidden
    size does not necessarily translate to a better performance.
    % Our
    % reruns use an improved implementation from the original authors, where
    % batch norm layers are removed. We fix the width to 32, selecting the top
    % 16 modes, fix batch size to 4, and train for 50 epochs.
    }

    \label{tab:results_elasticity}
    \begin{center}
    {\small
    \begin{tabularx}{\columnwidth}{Xrrrrrr}
      \toprule
       & \multirow{2}{*}{\makecell[r]{No. of \\ layers}}
       & \multirow{2}{*}{\makecell[r]{No. of \\ parameters}}
       & \multicolumn{3}{c}{N-MSE (\%)}
       & \multirow{2}{*}{\makecell[r]{Training \\ time (h)}} \\
      \cmidrule(r){4-6}
       & & & \makecell[r]{Mean}
       & \makecell[r]{Min}
       & \makecell[r]{Max}
       \\
      \midrule
    \multirow{3}{*}{geo-FNO (reproduced)}
     & 4 & 1,546,403 & 2.51 & 2.43 & 2.59 &  0.4 \\
     & 8 & 2,730,659 & 3.30 & 2.32 & 5.10 &  0.5 \\
     & 12 & 3,914,915 & 16.76 & 16.17 & 17.72 &  0.7 \\
     \midrule
     \multirow{6}{*}{F-FNO}
      & 4 & 3,205,763 & 3.16 & 2.23 & 4.98 &  1 \\
      & 8 & 4,338,051 & 2.05 & 2.04 & 2.06 &  1 \\
      & 12 & 5,470,339 & 1.96 & 1.93 & 1.98 &  2 \\
      & 16 & 6,602,627 & 1.86 & 1.83 & 1.88 &  2 \\
      & 20 & 7,734,915 & 1.84 & 1.82 & 1.86 &  2 \\
      & 24 & 8,867,203 & 1.74 & 1.70 & 1.78 &  2 \\
    \midrule
    \multirow{6}{*}{F-FNO-WS (F-FNO with weight sharing)}
     & 4 & 2,681,475 & 3.55 & 2.36 & 5.84 &  0.2 \\
     & 8 & 2,765,187 & 2.23 & 2.18 & 2.29 &  0.2 \\
     & 12 & 2,848,899 & 2.12 & 2.10 & 2.16 &  0.3 \\
     & 16 & 2,932,611 & 2.08 & 2.06 & 2.10 &  0.4 \\
     & 20 & 3,016,323 & 2.04 & 1.99 & 2.07 &  0.5 \\
     & 24 & 3,100,035 & 1.97 & 1.94 & 2.01 &  0.6 \\
      \bottomrule
    \end{tabularx}
    }
  \end{center}
  \end{table}

\begin{table}[h]
    \caption{Detailed performance on \Plasticity. These results are more detailed version of
    \cref{tab:meshes} in the main paper.
    We run three trials for each experiment, each with a different random seed.
    We report the mean N-MSE from the three trials, along with the min and max value.
    % Our
    % reruns use an improved implementation from the original authors, where
    % batch norm layers are removed. We fix the width to 32, selecting the top
    % 16 modes, fix batch size to 4, and train for 50 epochs.
    }

    \label{tab:results_plasticity}
    \begin{center}
    {\small
    \begin{tabularx}{\columnwidth}{Xrrrrrr}
      \toprule
       & \multirow{2}{*}{\makecell[r]{No. of \\ layers}}
       & \multirow{2}{*}{\makecell[r]{No. of \\ parameters}}
       & \multicolumn{3}{c}{N-MSE (\%)}
       & \multirow{2}{*}{\makecell[r]{Training \\ time (h)}} \\
      \cmidrule(r){4-6}
       & & & \makecell[r]{Mean}
       & \makecell[r]{Min}
       & \makecell[r]{Max}
       \\
      \midrule
    \multirow{3}{*}{geo-FNO (reproduced)}
     & 4 & 18,883,492 & 0.74 & 0.73 & 0.75 &  2 \\
     & 8 & 37,762,084 & 0.57 & 0.55 & 0.63 &  4 \\
     & 12 & 56,640,676 & 0.45 & 0.41 & 0.49 &  5 \\
     \midrule
     \multirow{6}{*}{F-FNO}
      & 4 & 1,846,920 & 0.48 & 0.47 & 0.51 &  4 \\
      & 8 & 3,684,488 & 0.32 & 0.31 & 0.34 &  8 \\
      & 12 & 5,522,056 & 0.25 & 0.24 & 0.26 &  12 \\
      & 16 & 7,359,624 & 0.22 & 0.21 & 0.22 &  16 \\
      & 20 & 9,197,192 & 0.20 & 0.18 & 0.22 &  20 \\
      & 24 & 11,034,760 & 0.18 & 0.17 & 0.18 &  24 \\
    \midrule
    \multirow{6}{*}{F-FNO-WS (F-FNO with weight sharing)}
     & 4 & 568,968 & 0.58 & 0.57 & 0.60 &  4 \\
     & 8 & 702,600 & 0.50 & 0.46 & 0.52 &  8 \\
     & 12 & 836,232 & 0.44 & 0.42 & 0.48 &  12 \\
     & 16 & 969,864 & 0.40 & 0.36 & 0.44 &  16 \\
     & 20 & 1,103,496 & 0.34 & 0.31 & 0.37 &  19 \\
     & 24 & 1,237,128 & 0.30 & 0.28 & 0.35 &  21 \\
      \bottomrule
    \end{tabularx}
    }
  \end{center}
  \end{table}

\begin{table}[h]
    \caption{The \ac{F-FNO} is flexible in its input representation.
    We find that different datasets benefit from having different
    features. Shown here is the optimal input combination for
    each dataset on the torus.}
    \begin{tabularx}{\linewidth}{Xrrrrrr}
    \toprule
    Dataset  & \rot{Vorticity} & \rot{Velocity} & \rot{Coordinates}
             & \rot{Viscosity} & \rot{Forcing}  & \\
    \midrule
    \TorusZongyi & \cmark & & \cmark \\
    \TorusKochkov & \cmark & \cmark & \cmark\\
    \TorusV & \cmark & & \cmark & \cmark \\
    \TorusVF & \cmark & & \cmark & \cmark & \cmark \\
    \bottomrule
  \end{tabularx}
  \label{tab:reps}
  \end{table}

% \begin{figure}[p]
%   \centering
%   \includegraphics[width=\linewidth]{figures/io_torus_kochkov}
%   \caption{On all the grid datasets -- \TorusZongyi, \TorusV, \TorusVF, and
%   \TorusKochkov, the goal is to evolve the vorticity on a surface of a 2D torus
%   over time. The input of the model (left) is the vorticity at a time step $t$
%   (potentially along with other contextual information such as the external
%   forces and the viscosity), and the output of the model (right) is the
%   vorticity at the next time step $t + 1$.}
%   \label{fig:io_torus}
% \end{figure}

\begin{figure}[p]
  \centering
  \includegraphics[width=\linewidth]{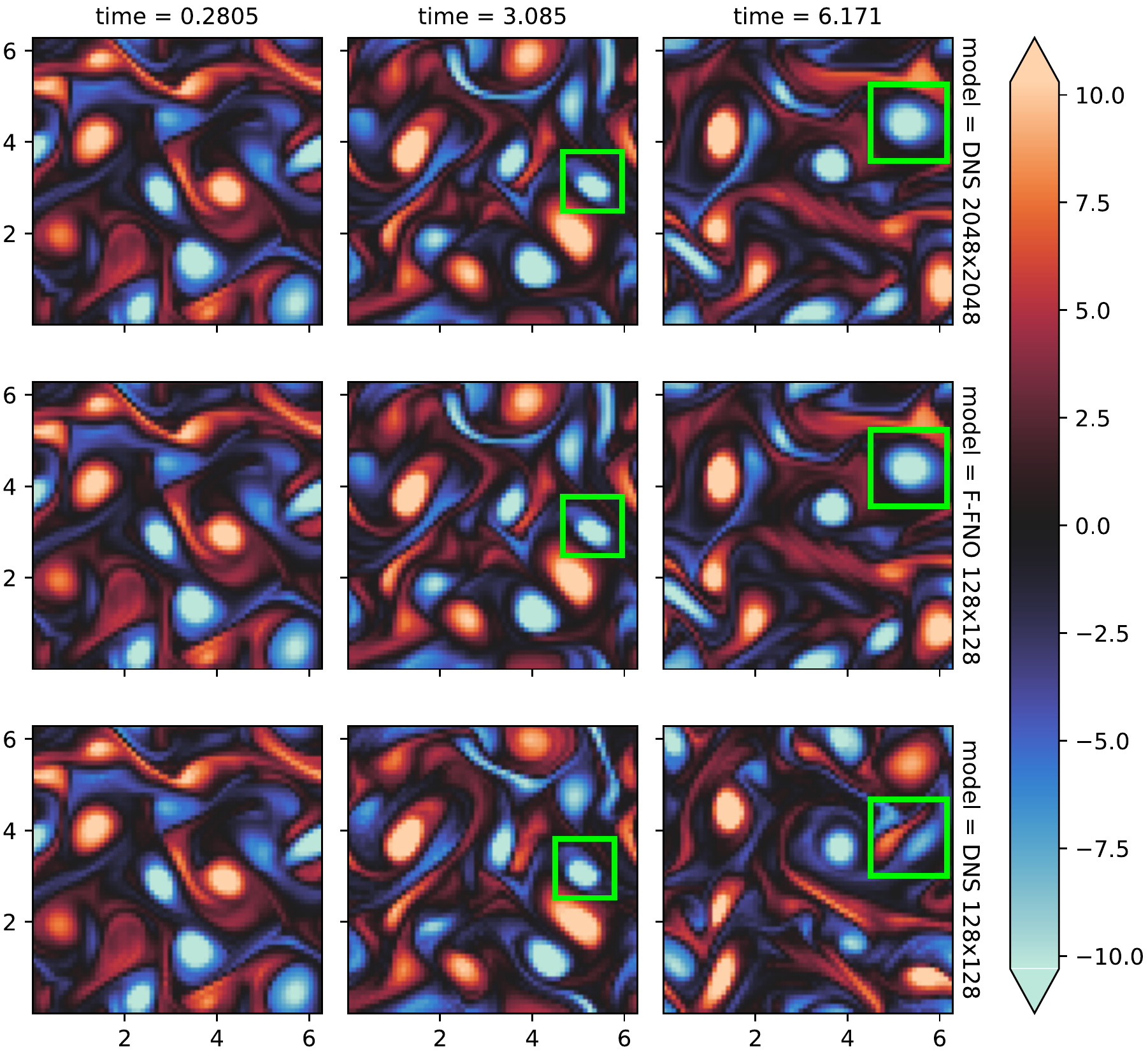}
  \caption{Similar to \citet{Kochkove2021Machine}, we visualize how the
  correlation with the ground truths varies between different models. The
  heatmaps represent the surface of a torus mapped onto a 2D grid, with color
  representing the vorticity (the spinning motion) of the fluid. We observe
  that the vorticity fields predicted by the \ac{F-FNO} trained on 128x128
  grids (middle row) correlates with the ground truths (top row) for longer
  than if we run \ac{DNS} on the same spatial resolution (bottom row). This is
  especially evident after 6 seconds of simulation time (compare the green
  boxes). In other words, for the same desired accuracy, the F-FNO requires a
  smaller grid input than a numerical solver. This observation is also backed
  up by \cref{fig:pareto}.}
  \label{fig:flows}
\end{figure}

% \begin{figure}[p]
%   \centering
%   \includegraphics[width=\linewidth]{figures/io_elasticity}
%   \caption{On \Elasticity, the input to the model (left) are the $(x,y)$ coordinates
%   of a point cloud, and the output of the model (right) is the stress value
%   on each point.}
%   \label{fig:io_elasticity}
% \end{figure}

% \begin{figure}[p]
%   \centering
%   \includegraphics[width=\linewidth]{figures/io_airfoil}
%   \caption{On \Airfoil, the input of the model (left) is the mesh grid, while
%   the output (right) is the Mach number (related to the speed) on each mesh
%   point.}
%   \label{fig:io_airfoil}
% \end{figure}

% \begin{figure}[p]
%   \centering
%   \includegraphics[width=\linewidth]{figures/io_plasticity}
%   \caption{On \Plasticity, the input of the model (left) is the 1D die function
%   (i.e., the boundary condition), while the output (right) is the displacement
%   of each point on a 2D over 20 time steps.}
%   \label{fig:io_plasticity}
% \end{figure}

\end{document}